\documentclass{article} 
\usepackage[final]{colm2026_conference}

\usepackage{microtype}
\usepackage{hyperref}
\usepackage{url}
\usepackage{booktabs}

\usepackage{lineno}

\usepackage{soul}
\usepackage{enumitem}
\usepackage{tabularx}
\usepackage{multirow}
\usepackage[dvipsnames]{xcolor}
\usepackage[most]{tcolorbox}
\usepackage{amssymb}
\usepackage{bbm}
\usepackage{colortbl}
\usepackage{makecell}
\usepackage{graphicx}
\usepackage[capitalize]{cleveref}
\usepackage{array}
\usepackage{fontawesome5}

\crefname{figure}{Figure}{Figures}
\crefname{table}{Table}{Tables}
\crefname{section}{Section}{Sections}
\crefname{equation}{Equation}{Equations}
\crefname{appendix}{Appendix}{Appendices}
\crefname{algorithm}{Algorithm}{Algorithms}
\definecolor{darkblue}{rgb}{0, 0, 0.5}
\hypersetup{colorlinks=true, citecolor=darkblue, linkcolor=darkblue, urlcolor=darkblue}

\renewcommand{\cite}{\citep}

\definecolor{deepgreen}{rgb}{0.0, 0.5, 0.0}

\title{Why Vision Fails as a Universal Bridge: Rectifying Modality Asynchrony in Multilingual MLLMs}

\author{
Yihang Du$^{1,2,*}$ \quad Juhao Liang$^{1,*}$ \quad Zhengzhao Lai$^{1}$ \quad Siyu Li$^{1}$ \quad Yan Hu$^{1,3,\dag}$  \\
$^{1}$The Chinese University of Hong Kong, Shenzhen \\
$^{2}$Shenzhen Loop Area Institute \\
$^{3}$National Health Data Institute (Shenzhen) \\
\texttt{yihangdu@slai.edu.cn}, \texttt{juhaoliang1@link.cuhk.edu.cn}, \texttt{huyan@cuhk.edu.cn}\\
$^{*}$ Equal contribution,  $^{\dag}$ Corresponding author
}

\begin{document}

\ifcolmsubmission
\linenumbers
\fi

\maketitle

\begin{abstract}
Multimodal large language models (MLLMs) exhibit substantial performance degradation in non-English visual reasoning, despite the strong multilingual competence of their text-only backbones. While mechanistic evidence from text-only models suggests that non-English inputs are routed through an English-centric latent space, the multimodal implications of this phenomenon remain unexplored. Through layer-wise mechanistic analysis, we identify the \textbf{Ghost Anchor} phenomenon: a temporal modality asynchrony where linguistic representations largely converge toward the English semantic manifold in early layers, while visual semanticization remains immature. Consequently, visual signals are physically present yet exert limited influence during the early alignment window. To address this, we propose \textbf{ANCHOR}, a training framework employing Proactive Visual Anchoring (PVA) to accelerate early visual semantic emergence and encourage visual representations to guide linguistic translation. Mechanistic interventions indicate that ANCHOR increases the influence of visual signals during early translation. Furthermore, experiments on xMMMU, MaXM, and CVQA show that ANCHOR improves aggregate performance over standard baselines across both fine-tuned and zero-shot languages.
\end{abstract}

\section{Introduction}
\label{intro}
Multimodal Large Language Models (MLLMs) have significantly advanced vision-language understanding by integrating LLMs with visual perception~\cite{tong2026beyond, bai2025qwen3}. However, while MLLMs excel in English, achieving equitable understanding across diverse languages remains a fundamental challenge~\cite{Pangea,xGQA}. Empirical evaluations reveal a substantial performance degradation in non-English contexts, indicating that current multimodal alignment remains highly English-dependent and fails to ground visual concepts within a truly language-agnostic space~\cite{zhong2024beyond,changpinyo2023maxm}.

A prevailing intuition attributes this disparity to data scarcity, motivating efforts to curate massive multilingual multimodal datasets~\cite{AyaVision,Pangea}. Yet, this data-centric approach incurs prohibitive costs and overlooks the internal representational mechanisms. Instead of forcing alignment externally, an emerging consensus acknowledges that MLLMs inherently develop a shared semantic space where diverse languages converge~\cite{zeng2025converging}. Ideally, heterogeneous inputs would converge toward a language-neutral semantic space through synchronized cross-modal interactions. In contemporary MLLMs, however, this shared space often empirically manifests as English-dominated because of language-centric architectures and imbalanced pre-training data~\cite{semantic-hub}. Within this empirically observed geometry, visual signals---grounded in physical reality---should act as active anchors that ground representations across linguistic boundaries~\cite{LRM-LLaVA,M3P,Uc2}.

\begin{figure*}[t]
    \centering
    \includegraphics[width=\linewidth]{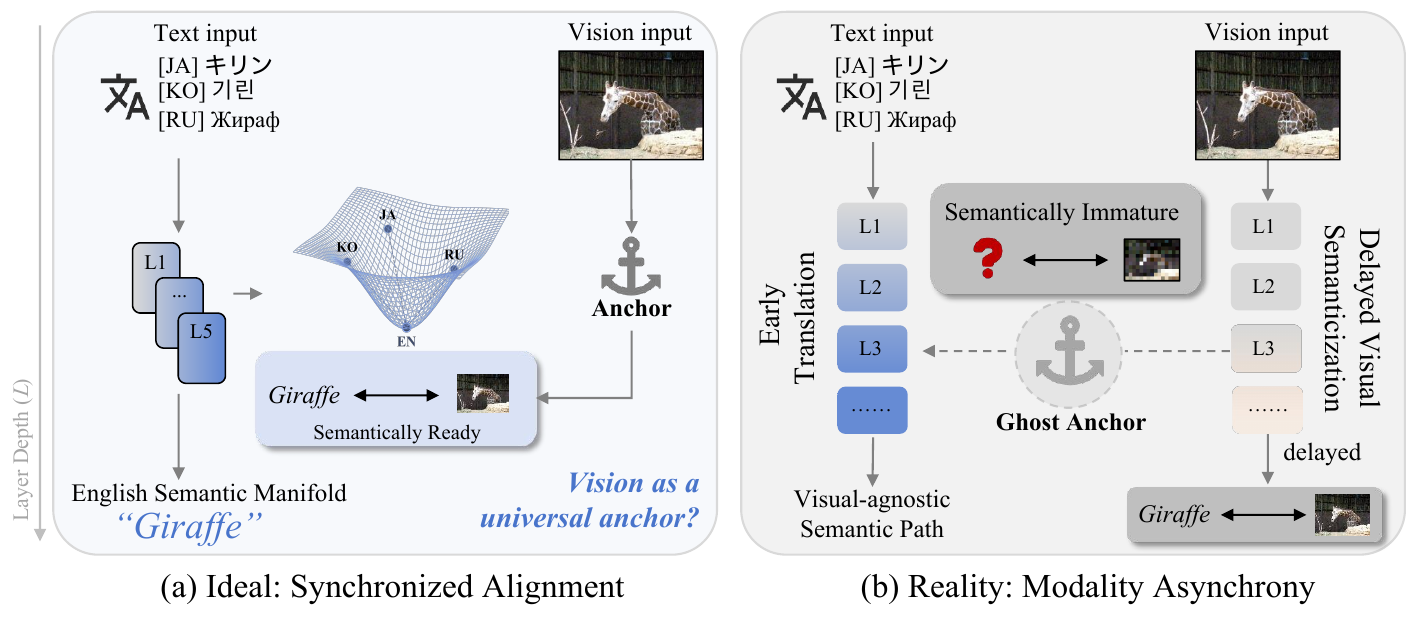}
    \caption{\textbf{The Ghost Anchor phenomenon.} (a) Ideal synchronized cross-modal alignment. (b) The reality of modality asynchrony: rapid linguistic translation bypasses delayed visual semantic emergence in early layers.}
    \label{fig:fig1}
\end{figure*}

However, the empirical reality is far from this synchronized ideal. Despite the theoretical promise of vision as a universal anchor, visual signals provide limited assistance for cross-lingual alignment: non-English languages continue to exhibit systematic deficits even when visual grounding should theoretically facilitate comprehension~\cite{bugliarello2022iglue,zheng2025mma}. This disconnect raises a critical question: \textit{Why does the visual modality fail to serve as the anticipated bridge?}

We hypothesize that the answer lies in the temporal dynamics of the Transformer architecture. Through a rigorous layer-wise mechanistic analysis, we uncover a profound \textbf{Modality Asynchrony}: linguistic representations exhibit substantial convergence toward the English semantic manifold in early layers, while visual grounding remains semantically immature until much deeper in the network. As illustrated in \cref{fig:fig1}, this early linguistic convergence emerges well before visual semantics become available. Consequently, the linguistic stream is forced to map to the English space ``blindly'', bypassing visual guidance.

Modality Asynchrony creates what we term the \textbf{Ghost Anchor} phenomenon: the visual modality is physically present in the computation yet remains semantically invisible during the critical early alignment phase (\cref{fig:fig1}). Consequently, the model defaults to a visual-agnostic translation path, which fundamentally degrades its non-English visual reasoning performance. 

To rectify this asynchrony, we propose \textbf{ANCHOR}, a training framework designed to synchronize dual-stream convergence. By accelerating visual semantic emergence through Proactive Visual Anchoring (PVA), ANCHOR encourages visual signals to become grounded early and influence the linguistic translation process. Experiments on benchmarks such as xMMMU~\cite{Pangea}, MaXM~\cite{changpinyo2023maxm}, and CVQA~\cite{romero2024cvqa} show that our approach improves cross-lingual visual performance.

In this work, we make the following principal contributions:
\begin{itemize}[leftmargin=1.5em, nosep, itemsep=0.5ex]
    \item We systematically investigate the internal dynamics of multilingual MLLMs and formalize the \textbf{Ghost Anchor} phenomenon, where \textbf{Modality Asynchrony} renders visual signals semantically inaccessible during early alignment.
    \item We provide mechanistic evidence through layer-wise analysis and controlled interventions, showing that visual information exerts negligible causal influence on the linguistic translation trajectory during the early alignment window.
    \item Building on these insights, we propose \textbf{ANCHOR}, a training framework that rectifies asynchrony by accelerating visual semantic emergence through Proactive Visual Anchoring (PVA), which explicitly supervises early-layer visual representations using an external visual foundation model.
\end{itemize}

\section{Related Work}
\label{sec:related work}
\paragraph{Mechanistic Perspectives on Cross-lingual Alignment}

Cross-lingual alignment has been primarily studied in text-only multilingual models, where semantically equivalent expressions gradually converge in hidden space and support transfer~\cite{zeng2025converging,zhong2024beyond}. Using logit-lens and causal-tracing style probes, mechanistic analyses further show that many multilingual LLMs process non-English inputs through an English-centric internal representation space, where non-English tokens are implicitly translated into an English semantic pivot before being mapped to the target output~\cite{wendler-etal-2024-llamas,schut2025multilingual,zeng2025converging}. At the same time, strong global alignment can hurt language-specific competence, especially for low-resource or distant languages~\cite{elshabrawy2025alignment}. This tension is still analyzed almost entirely without vision, leaving unclear how visual signals reshape (or fail to reshape) cross-lingual alignment inside MLLMs~\cite{semantic-hub,sundar2025steering}. Our work fills this gap by extending mechanistic analysis to multimodal settings and showing that visual signals can fail exactly when modality asynchrony delays visual semantic emergence during the critical alignment window.

\paragraph{Multilingual Large Vision Language Models}
Early multilingual vision-language models such as M$^3$P~\cite{M3P}, UC$^2$~\cite{Uc2}, and related BERT-era frameworks~\cite{huang2021multilingual} explicitly optimized cross-lingual and cross-modal objectives during pretraining. In the LLM era, mainstream multilingual MLLMs rely more on scale-first pipelines, where large translated or synthetic instruction mixtures are expected to induce multilingual visual competence implicitly~\cite{Pangea,AyaVision}. This data-centric strategy improves benchmarks but incurs high annotation cost and offers limited control over how multilingual text and visual semantics are aligned internally. More targeted methods add alignment modules or objectives, including mBLIP~\cite{geigle-etal-2024-mblip}, LRM-LLaVA~\cite{LRM-LLaVA}, and M$^2$-VLP~\cite{10.1145/3696410.3714861}. However, their supervision is still mainly applied at output or global-distribution levels rather than explicitly constraining layer-wise cross-modal interaction during the critical alignment phase. In contrast, our approach starts from a mechanistic diagnosis of modality asynchrony and directly targets the critical layer-wise interaction window.

\section{Unveiling Modality Asynchrony: A Mechanistic Analysis}

\textbf{Do visual signals truly function as language-agnostic anchors in multilingual MLLMs?} To answer this, we systematically investigate both the independent evolutionary trajectories of multilingual text and visual signals, and how they interact within the model's internal representations.

\paragraph{The English Semantic Manifold} To perform joint reasoning, MLLMs inherently project heterogeneous inputs (e.g., multilingual text and visual signals) into a shared representational space~\cite{song2025bridge}. During inference, representations from both diverse languages~\cite{zeng2025converging,zhao2024large} and cross-modal visual inputs~\cite{semantic-hub,venhoff2025visual} progressively converge into a unified geometry. Driven by the language-centric architecture of contemporary MLLMs~\cite{shen2025vl,shu2025large} and imbalanced pre-training data, this shared hub pragmatically manifests as an English-dominated space~\cite{schut2025multilingual,zhong2024beyond}. Thus, we formalize the English Semantic Manifold ($\mathcal{S}_{en}$) as the target coordinate system where intermediate hidden states $h_l$ progressively converge across layers: $h_{l} \xrightarrow{\text{layers}} h^{*} \in \mathcal{S}_{en}$.

\paragraph{Dynamics of Multilingual and Modal Convergence.} 
Let $\mathcal{S}_{src}$ denote the source-language representational space and $\mathcal{S}_{vis}$ the initial perceptual space occupied by visual tokens. Within this unified framework, the recurrence $h_l=f_l(h_{l-1})$ traces an incremental trajectory toward the English semantic manifold, where $h_l$ denotes the hidden state at layer $l$. We use $F_l=f_l\circ\cdots\circ f_1$ to denote the cumulative transformation through layer $l$. Both streams exhibit convergence toward $\mathcal{S}_{en}$, yet follow distinct trajectories: the linguistic stream is hypothesized to undergo an implicit transition from $\mathcal{S}_{src}$ toward $\mathcal{S}_{en}$ under $F_l$, while the visual stream undergoes visual semanticization from $\mathcal{S}_{vis}$ toward $\mathcal{S}_{en}$.

Crucially, for the MLLM to perform true multimodal reasoning, the implicit translation of the text stream should theoretically be grounded in the visual context. Since Transformer architectures aggregate information via layer-wise self-attention, this grounding hinges on the \textbf{temporal coordination} between the two paths. If visual tokens have not yet achieved visual semanticization during the specific layer window where translation occurs, the linguistic stream could default to a purely text-based transformation. Such a mechanistic asynchrony would result in a bypass of visual evidence, potentially explaining the reasoning failures observed in multilingual contexts. In the following section, we empirically measure the precise layer-wise timing of these two processes to test this asynchrony hypothesis.

\subsection{Modality Asynchrony: Early Translation, Delayed Semanticization}
\label{sec:asynchrony}

\begin{figure}[t]
    \centering
    \includegraphics[width=\linewidth]{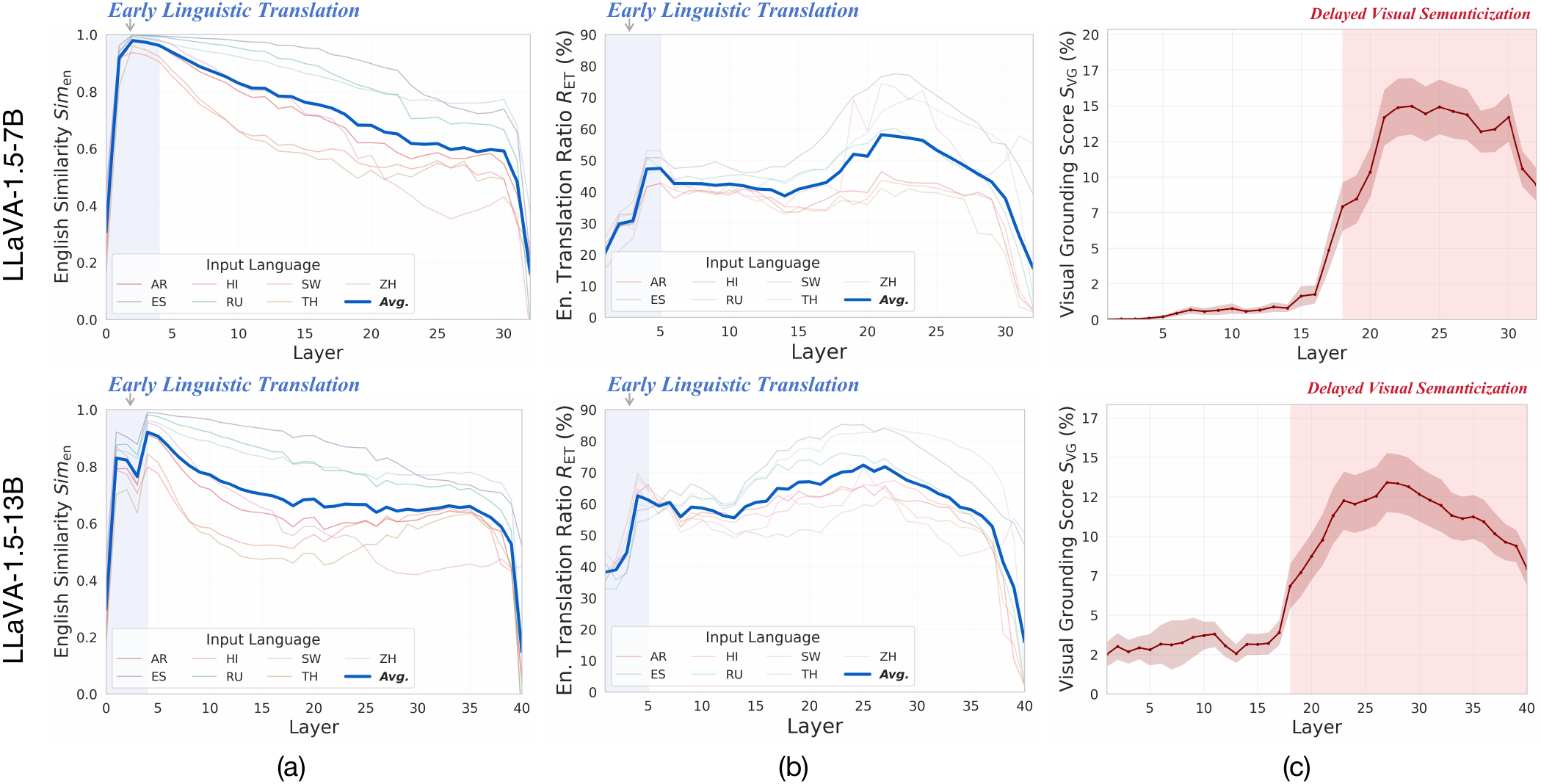}
    \caption{\textbf{Modality Asynchrony}. \textbf{(a-b)} Layer-wise English Similarity ($\text{Sim}_{en}$) and English Translation Ratio ($R_{ET}$) in text stream. \textbf{(c)} Visual Grounding Score ($S_{VG}$) in vision stream. Shaded regions indicate the standard error of the mean (SEM).}
    \label{fig:fig2}
\end{figure}

To investigate whether these pathways evolve synchronously, we extract intermediate hidden states $h_l$ under a controlled input condition where visual tokens are paired with multilingual text descriptions. All of the experimental details and metric definitions can be found in Appendix~\ref{sec:exp_setup}. We monitor the dual-stream convergence using three complementary metrics. All vocabulary-level projections employ the \textit{logit lens} technique~\cite{nostalgebraist2020interpreting}, which maps intermediate hidden states to probability distributions over the vocabulary via the pre-trained unembedding matrix.

\begin{itemize}[leftmargin=*]
    \item \textbf{Text Stream:} We track the continuous geometric rotation and discrete lexical transition toward $\mathcal{S}_{en}$ using \textbf{English Similarity ($\text{Sim}_{en}$)} and the \textbf{English Translation Ratio ($R_{ET}$)}, respectively. Specifically, $\text{Sim}_{en}$ measures the cosine similarity between the mean hidden states of the source and English reference sequences. $R_{ET}$ extracts the top-$k$ decoded words from each text token via the logit lens and measures the proportion classified as English by a zero-shot language classifier.
    \item \textbf{Vision Stream:} We directly probe the visual semanticization process using the \textbf{Visual Grounding Score ($S_{VG}$)}, tracking when visual tokens decode into ground-truth object labels via the logit lens.
\end{itemize}

\paragraph{Finding 1: Early Linguistic Translation.}
Both $\text{Sim}_{en}$ and $R_{ET}$ reveal rapid linguistic convergence within the initial layers. As shown in \cref{fig:fig2}(a-b), source language representations efficiently rotate toward the English representation and decode into English vocabulary as early as layers 1--5. This demonstrates that the English-dominant linguistic transition becomes strongly detectable in the earliest stages of the trajectory. 

\paragraph{Finding 2: Delayed Visual Semanticization.}
In contrast, visual semantics emerge later. \cref{fig:fig2}(c) shows that $S_{VG}$ remains near-zero throughout early-to-middle layers, only rising beyond layer 20. Bridging the cross-modal gap ($h_l : \mathcal{S}_{vis} \rightarrow \mathcal{S}_{en}$) appears more demanding than intra-lingual translation, creating a pronounced Modality Asynchrony.

\paragraph{The Ghost Anchor Phenomenon.}
The asynchrony suggests a mechanistic misalignment: by the time the linguistic stream settles into $\mathcal{S}_{en}$, visual tokens remain close to their initial perceptual manifold $\mathcal{S}_{vis}$. At this stage, they primarily encode low-level sensory features and provide limited assistance to the translation process. We characterize this temporal separation as the Ghost Anchor phenomenon---visual tokens are physically present but exert limited measurable influence during early alignment. We test this hypothesis through controlled intervention in \cref{sec:causal_validation}.

\subsection{Causal Validation of the Ghost Anchor}
\label{sec:causal_validation}

\begin{figure}[t]
    \centering
    \includegraphics[width=0.9\linewidth]{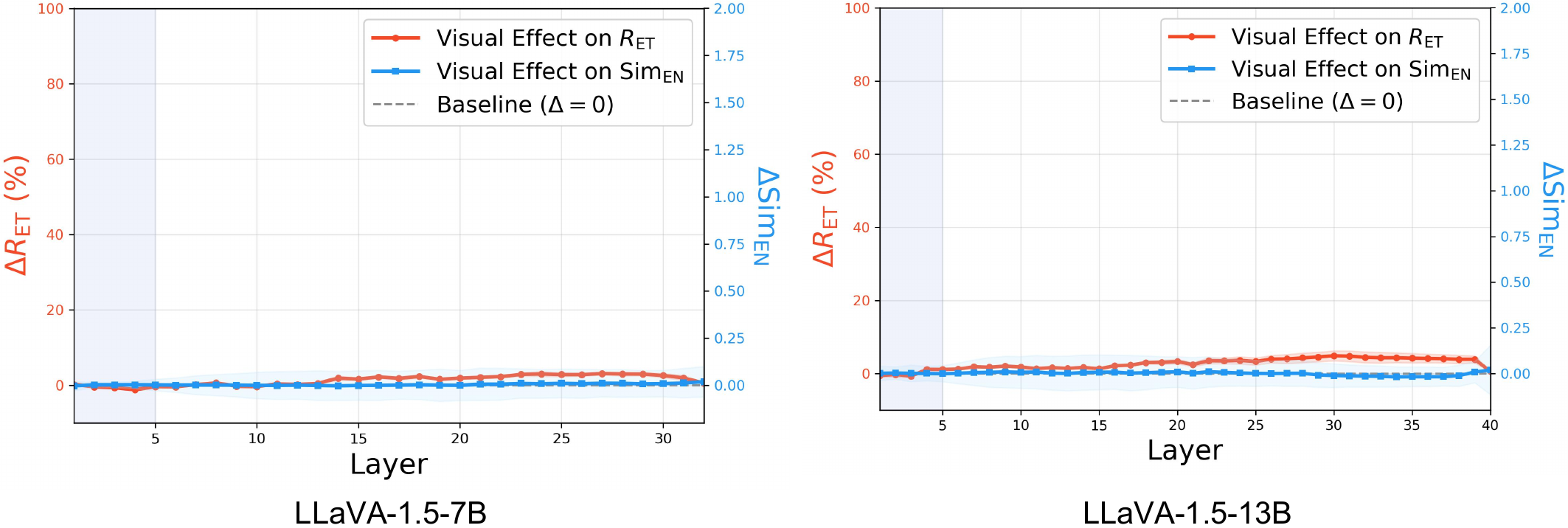}
    \caption{\textbf{Visual Causal Effect} (LLaVA-1.5-7B left, 13B right). Difference between Normal and Noise inputs ($\Delta = \text{Normal} - \text{Noise}$) across layers. Shaded regions indicate the standard error of the mean (SEM).}
    \label{fig:en_ratio_noise}
\end{figure}

While $R_{ET}$ is already high in early layers (\cref{fig:fig2}(b)), the critical question is whether this translation trajectory is sensitive to visual context or predominantly driven by textual priors. To test this, we replace each visual input with matched Gaussian noise---preserving the input dimensions, patch count, and token positions while substantially disrupting recognizable objects and scene structure (see Appendix~\ref{sup:noise_intervention}). We therefore measure the \textbf{Visual Causal Effect} on the text stream via both the continuous geometric shift ($\Delta \text{Sim}_{en}$) and the discrete lexical transition ($\Delta R_{ET}$). As shown in \cref{fig:en_ratio_noise}, both metrics remain near-zero across early layers for LLaVA-1.5. These small changes indicate that the measured early translation trajectory has limited sensitivity to high-level visual content under this intervention, providing evidence consistent with the Ghost Anchor phenomenon. We further conduct experiments on a broader range of SOTA MLLMs in Appendix~\ref{sec:appendix_sota}, which show similarly small effects.

\section{ANCHOR: Rectifying Modality Asynchrony through Proactive Visual Anchoring}
\label{sec:anchor}
\subsection{Method Overview}
Motivated by our mechanistic discovery of the Ghost Anchor effect, we propose ANCHOR to structurally synchronize the disparate evolutionary trajectories of text and vision. Unlike data-centric interventions that force external alignment, ANCHOR explicitly supervises the early-layer evolution of visual representations. Its core component, Proactive Visual Anchoring (PVA), leverages an external visual foundation model (VFM) to enforce visual semantic readiness before the critical alignment window closes, ensuring visual features actively guide linguistic translation.

\subsection{Proactive Visual Anchoring}
Standard vision-language training treats the vision encoder as a fixed feature extractor and only adapts the language side. However, under text-only supervision, fine-grained visual information may not be preserved as visual tokens propagate through the LLM's early layers. By the time the semantic verbalization process completes, the cross-modal alignment window has already closed. PVA therefore uses an external VFM as a semantic target to encourage visually grounded representations in the early layers.

 \paragraph{Alignment Objective.} Given the MLLM's intermediate hidden states $h_{i,l} \in \mathbb{R}^{d_{\mathrm{MLLM}}}$ for visual token $i$ at layer $l$, we pass them through a lightweight alignment network $\text{MLP}_{align}(\cdot)$ and compare the output against the semantic supervisor from a pre-trained VFM. Following the formulation in prior representation alignment work, the Proactive Visual Anchoring loss is defined as the negative mean cosine similarity across all visual tokens:
\begin{equation}
    \mathcal{L}_{anc} = -\frac{1}{|\mathcal{V}_{vis}|} \sum_{i \in \mathcal{V}_{vis}} \text{cos}\big(\text{MLP}_{align}(h_{i,L_{early}}), z_{i, VFM}\big),
    \label{eq:pva_loss}
\end{equation}
where $|\mathcal{V}_{vis}|$ denotes the number of visual tokens, and $z_{i, VFM} \in \mathbb{R}^{d_{\mathrm{VFM}}}$ is the fine-grained semantic feature extracted from SigLIP-SO400M/14~\cite{tschannen2025siglip} for visual token $i$. The alignment network $\text{MLP}_{align}: \mathbb{R}^{d_{\mathrm{MLLM}}} \rightarrow \mathbb{R}^{d_{\mathrm{VFM}}}$ consists of a two-layer perceptron with GELU activation that projects the MLLM hidden states into the VFM feature space.

The key design choices are: (1) we apply supervision at layer $L_{early}$, where visual semantics should become available for cross-modal alignment; and (2) we use a VFM (SigLIP-SO400M/14) as the supervisor because it provides visually grounded semantic features that can serve as a shared reference across target languages.

\textbf{Combined Training Objective.} The full training objective combines the standard visual instruction tuning loss with the PVA anchoring loss:
\begin{equation}
    \mathcal{L} = \mathcal{L}_{SFT} + \alpha \cdot \mathcal{L}_{anc},
    \label{eq:full_loss}
\end{equation}
where $\mathcal{L}_{SFT}$ is the standard next-token prediction loss on multilingual image-text pairs, and $\alpha$ controls the strength of the anchoring regularization. During training, we apply full-parameter fine-tuning (FFT) to both the multimodal projector and the first $L_{early}$ transformer layers, freeze the remaining layers and the vision encoder, and train the alignment network $\text{MLP}_{align}$. The ablation in \cref{sec:ablation} suggests that restricting the intervention to this early window better preserves downstream performance than unconstrained full-depth updates.

By supervising the early-layer trajectory of visual tokens, PVA effectively shifts the visual semantic emergence curve earlier in the model's depth, ensuring that when the linguistic stream begins its cross-modal alignment phase, grounded visual evidence is already available and can actively influence the alignment process.

\section{Empirical Results and Analysis}
To empirically validate the ANCHOR framework, we structure our evaluation around three core research questions: \textbf{RQ1} Multilingual Performance: Does ANCHOR improve visual reasoning across fine-tuned and zero-shot languages without degrading native English proficiency? \textbf{RQ2} Mechanistic Rectification: Does the framework mitigate the Ghost Anchor phenomenon by increasing the influence of visual signals during early translation? \textbf{RQ3} Intervention Design: Does explicit PVA supervision during the early alignment window improve performance over unanchored early-layer adaptation?

\subsection{Experiment Settings}

\paragraph{Implementation Details} We implement ANCHOR using LLaVA-1.5 as our base MLLM, a classic architecture that is widely used for interpretability research. For PVA, we employ SigLIP-SO400M/14 as the external Visual Foundation Model. Following our mechanistic analysis, we set $L_{early}=10$ as the early alignment window. Models are fine-tuned on 135K image-text pairs (15K × 9 languages) from ShareGPT4V~\cite{chen2024sharegpt4v}. See \cref{sec:train_dataset} for dataset details.

\paragraph{Baselines}
To isolate the efficacy of PVA, we compare ANCHOR against two baselines: (1) \textbf{Std. LoRA}, which fine-tunes without any visual-side intervention; and (2) \textbf{Early-Layer FFT}, which applies full-parameter fine-tuning to the first $L_{early}$ transformer layers but omits the PVA objective. 

\paragraph{Benchmarks}
We evaluate multilingual proficiency across three benchmarks targeting distinct dimensions: (1) \textbf{xMMMU}~\cite{Pangea} for deep semantic reasoning in expert domains; (2) \textbf{MaXM}~\cite{changpinyo2023maxm} for cross-lingual transfer on low-resource languages; and (3) \textbf{CVQA}~\cite{romero2024cvqa} for culturally nuanced visual mappings. Crucially, our evaluation covers both languages included in multilingual SFT and languages unseen during training, enabling us to assess generalization across both seen and zero-shot settings.
\colorlet{trainedbg}{pink!30} 
\colorlet{zerobg}{cyan!15} 

\newcolumntype{C}[1]{>{\centering\arraybackslash}p{#1}}

\begin{table*}[t]
\centering
\setlength{\aboverulesep}{0pt}
\setlength{\belowrulesep}{0pt}
\renewcommand{\arraystretch}{1.2}

\textbf{(a) Results on the xMMMU}
\vspace{2mm}
\resizebox{\textwidth}{!}{
\begin{tabular}{ll C{0.8cm} C{0.8cm}C{0.8cm}C{0.8cm}C{0.8cm} C{1.0cm}C{1.0cm} C{1.0cm}}
\toprule
\multirow{2}{*}{\textbf{Size}} & \multirow{2}{*}{\textbf{Models}} & \multirow{2}{*}{\textbf{en}} & \multicolumn{4}{c}{\cellcolor{trainedbg}\textbf{SFT Languages \faHotjar}} & \multicolumn{2}{c}{\cellcolor{zerobg}\textbf{Zero-shot \faSnowflake}} & \multirow{2}{*}{\textbf{Avg.}} \\
\cmidrule(lr{2pt}){4-7} \cmidrule(lr{2pt}){8-9} 
& & & \cellcolor{trainedbg}\textbf{ar} & \cellcolor{trainedbg}\textbf{fr} & \cellcolor{trainedbg}\textbf{hi} & \cellcolor{trainedbg}\textbf{pt} & \cellcolor{zerobg}\textbf{id} & \cellcolor{zerobg}\textbf{ja} & \\
\midrule
\multirow{4}{*}{7B}  
 & LLaVA-1.5          & \textbf{35.2} & 29.9 & 34.6 & 28.2 & 33.7 & 31.6 & 32.0 & 32.2 \\
 & \quad + Std. LoRA       & 34.9 & 32.9 & 36.6 & 30.2 & 36.0 & 32.0 & 30.9 & 33.4 \\ 
  & \quad + Early-Layer FFT & 32.6 & 28.9 & 32.2 & 24.1 & 31.6 & \textbf{34.3} & 32.0 & 30.8      \\ 
\rowcolor{gray!15} & \textbf{\quad + ANCHOR}  & 35.0 & \textbf{34.8} & \textbf{39.1} & \textbf{32.6} & \textbf{39.2} & 33.9  & \textbf{34.4} & \textbf{35.6}  \\ 
\midrule
\multirow{4}{*}{13B} 
 & LLaVA-1.5               & 34.8 & 30.9 & 35.6 & 29.2 & 37.0 & 33.7 & 30.9 & 33.2 \\
 & \quad + Std. LoRA       & 34.8 & 31.5 & 35.6 & 30.2 & 36.0 & 33.0 & 31.2 & 33.2 \\ 
 & \quad + Early-Layer FFT  & 33.2 & 31.2 & 35.2 & 29.6 & 33.7 & 30.6 & 28.3 & 31.7   \\ 
\rowcolor{gray!15} & \textbf{\quad + ANCHOR}  & \textbf{35.1}  & \textbf{36.3} & \textbf{37.5} & \textbf{34.2} & \textbf{37.3} & \textbf{35.2} & \textbf{33.1} & \textbf{35.5} \\ 
\bottomrule
\end{tabular}
}
\vspace{2mm}
\textbf{(b) Results on the MaXM}
\vspace{2mm}
\resizebox{\textwidth}{!}{
\begin{tabular}{ll C{0.8cm} C{0.8cm}C{0.8cm}C{0.8cm}C{0.8cm} C{1.0cm}C{1.0cm} C{1.0cm}}
\toprule
\multirow{2}{*}{\textbf{Size}} & \multirow{2}{*}{\textbf{Models}} & \multirow{2}{*}{\textbf{en}} & \multicolumn{4}{c}{\cellcolor{trainedbg}\textbf{SFT Languages \faHotjar}} & \multicolumn{2}{c}{\cellcolor{zerobg}\textbf{Zero-shot \faSnowflake}} & \multirow{2}{*}{\textbf{Avg.}} \\
\cmidrule(lr{2pt}){4-7} \cmidrule(lr{2pt}){8-9} 
& & & \cellcolor{trainedbg}\textbf{fr} & \cellcolor{trainedbg}\textbf{hi} & \cellcolor{trainedbg}\textbf{th} & \cellcolor{trainedbg}\textbf{zh} & \cellcolor{zerobg}\textbf{he} & \cellcolor{zerobg}\textbf{ro} & \\
\midrule
\multirow{4}{*}{7B}  
 & LLaVA-1.5          & 49.4 & 32.6 & 17.7 & 17.2 & 27.8 & 12.9 & 15.1 & 24.7 \\
 & \quad + Std. LoRA  & 49.0 & 33.0 & 37.7 & 30.2 & 30.3 & 16.1 & \textbf{18.3} & 30.7  \\ 
 & \quad + Early-Layer FFT & 40.5 & 30.7 & 32.3 & 26.9 & \textbf{36.5} & \textbf{19.3} & 16.9 & 29.0      \\ 
\rowcolor{gray!15} & \textbf{\quad + ANCHOR}  & \textbf{51.3} & \textbf{36.8} & \textbf{38.4} & \textbf{38.2} & 29.3  & 14.7 & \textbf{18.3} &  \textbf{32.4} \\ 
\midrule
\multirow{4}{*}{13B} 
 & LLaVA-1.5          & 51.8 & 30.3 & 15.8 & 20.1 & 22.0 & 14.3 & 23.6 & 25.4 \\
 & \quad + Std. LoRA       & 48.6 & 34.9 & 38.3 & 32.8 & 30.3 & 16.1 & 26.8 & 32.5 \\ 
  & \quad + Early-Layer FFT & 36.6 & 35.2 & 33.5 & 34.0 & \textbf{40.0} & 13.6 & \textbf{35.6} &  32.6      \\ 
\rowcolor{gray!15} & \textbf{\quad + ANCHOR}  & \textbf{54.1} & \textbf{39.1} &  \textbf{39.3} & \textbf{37.2} & 36.9 & \textbf{20.5} & 28.1 & \textbf{36.5} \\ 
\bottomrule
\end{tabular}
}
\caption{Performance comparison on multilingual multimodal benchmarks: (a) xMMMU  and (b) MaXM. Best results are highlighted in \textbf{bold}.}
\label{tab:main_results_merged}
\end{table*}

\subsection{Quantitative Results}
\label{sec:quantitative}

To systematically assess the macroscopic generalization of our framework, we evaluate ANCHOR on xMMMU, MaXM, and CVQA. Results across all evaluations consistently exhibit three core trends that corroborate our hypotheses:

\textbf{Retention of the English Semantic Manifold.} A persistent challenge in cross-lingual alignment is degradation of the backbone's native English proficiency. Our evaluations indicate that ANCHOR generally preserves---and in certain cases, enhances---this capability. For instance, while maintaining stable English performance on xMMMU, ANCHOR yields improvements on MaXM. These results suggest that early-layer visual grounding via PVA can integrate visual anchors without substantially interfering with the LLM's established English-centric reasoning pathways.

\textbf{Substantial Gains in Target SFT Languages.} For non-English languages explicitly encountered during fine-tuning, ANCHOR demonstrates pronounced improvements. Compared with the degradation observed for the Early-Layer FFT baseline, ANCHOR better preserves performance while updating early layers. For example, on the 13B CVQA benchmark, standard early-layer adaptation reduces Arabic accuracy from 38.4\% to 28.1\%; in contrast, ANCHOR improves it to 40.1\% and raises the average across the six displayed language--region pairs to 50.3\%, surpassing both the base model and standard LoRA on this displayed subset. This contrast supports our mechanistic premise and indicates that explicit visual anchoring is beneficial when adapting early layers.

\textbf{Robust Zero-Shot Cross-Lingual Generalization.} The benefits of mitigating modality asynchrony extend to unseen linguistic spaces, reducing the cross-lingual "alignment tax" often incurred during SFT. When transferring to zero-shot languages, standard baselines can degrade because of overfitting to the target SFT distribution. ANCHOR cushions this degradation on CVQA and achieves absolute zero-shot improvements on xMMMU (e.g., zero-shot Japanese improving from 32.0\% to 34.4\% on the 7B model). These results suggest that mitigating the Ghost Anchor phenomenon can produce a more transferable cross-modal alignment space.

\colorlet{trainedbg}{pink!30} 
\colorlet{zerobg}{cyan!15} 

\begin{table*}[t] 
\centering
\small 
\setlength{\tabcolsep}{4.5pt} 
\setlength{\aboverulesep}{0pt}
\setlength{\belowrulesep}{0pt}
\renewcommand{\arraystretch}{1.2}

\resizebox{\textwidth}{!}{
\begin{tabular}{ll ccc | ccc | c}
\toprule
\multirow{2}{*}{\textbf{Size}} & \multirow{2}{*}{\textbf{Models}} & \multicolumn{3}{c|}{\cellcolor{trainedbg}\textbf{SFT Languages \faHotjar}} & \multicolumn{3}{c|}{\cellcolor{zerobg}\textbf{Zero-shot \faSnowflake}} & \multirow{2}{*}{\textbf{Avg.}} \\
\cmidrule(lr{2pt}){3-5} \cmidrule(lr{2pt}){6-8} 
& & \cellcolor{trainedbg}\textbf{ar-EG} & \cellcolor{trainedbg}\textbf{es-ES} & \cellcolor{trainedbg}\textbf{zh-CN} & \cellcolor{zerobg}\textbf{am-ET} & \cellcolor{zerobg}\textbf{ja-JP} & \cellcolor{zerobg}\textbf{ro-RO} & \\
\midrule
\multirow{4}{*}{7B} 
 & LLaVA-1.5                & 32.5 & 67.0 & 48.6 & 26.5 & 35.0 & \textbf{51.0} & \textbf{43.4} \\
 & \quad + Std. LoRA        & 31.5 & 62.0 & 44.7 & 26.1 & 35.5 & 46.7 & 41.1 \\ 
 & \quad + Early-Layer FFT  & 29.6 & 50.3 & 42.8 & \textbf{27.4} & 35.5 & 44.4 & 38.3 \\ 
\rowcolor{gray!15} & \textbf{\quad + ANCHOR}  & \textbf{37.3} & \textbf{69.3} & \textbf{49.1} & 26.4 & \textbf{36.9} & 37.4 & 42.7 \\
\midrule
\multirow{4}{*}{13B} 
 & LLaVA-1.5                & 38.4 & 70.4 & 54.3 & 25.2 & 45.8 & 57.0 & 48.5 \\
 & \quad + Std. LoRA   & 36.0 & 67.9 & 50.5 & 26.1 & 42.4 & 52.3 & 45.9 \\ 
 & \quad + Early-Layer FFT  & 28.1 & 52.2 & 41.8 & \textbf{36.8} & 37.0 & 44.7 & 40.1 \\ 
\rowcolor{gray!15} & \textbf{\quad + ANCHOR}  & \textbf{40.1} & \textbf{70.8} & \textbf{59.4} & 26.0 & \textbf{47.5} & \textbf{58.0} & \textbf{50.3} \\ 
\bottomrule
\end{tabular}
}
\caption{Performance on CVQA. We show six language--region pairs whose languages include both those used in SFT and those unseen during training; full results are reported in \cref{tab:CVQA_local_full_reordered}. The Avg. column reports the average accuracy across the six displayed language--region pairs. Best results are highlighted in \textbf{bold}.}
\label{tab:CVQA_local_summary}
\end{table*}

\paragraph{Generalization to Newer MLLMs.}
To examine whether ANCHOR extends beyond LLaVA-1.5, we additionally evaluate it on LLaVA-NeXT (7B/13B) and InternVL3 (8B/14B), using the same training and evaluation settings as the corresponding Std. LoRA baselines. ANCHOR consistently improves over Std. LoRA across the English, SFT-language, and zero-shot-language groups. Detailed gains on xMMMU, MaXM, and CVQA are reported in \cref{tab:newer_mllm_xmmmu,tab:newer_mllm_maxm,tab:newer_mllm_cvqa}.

\begin{figure}[!h]
    \centering
    \includegraphics[width=\linewidth]{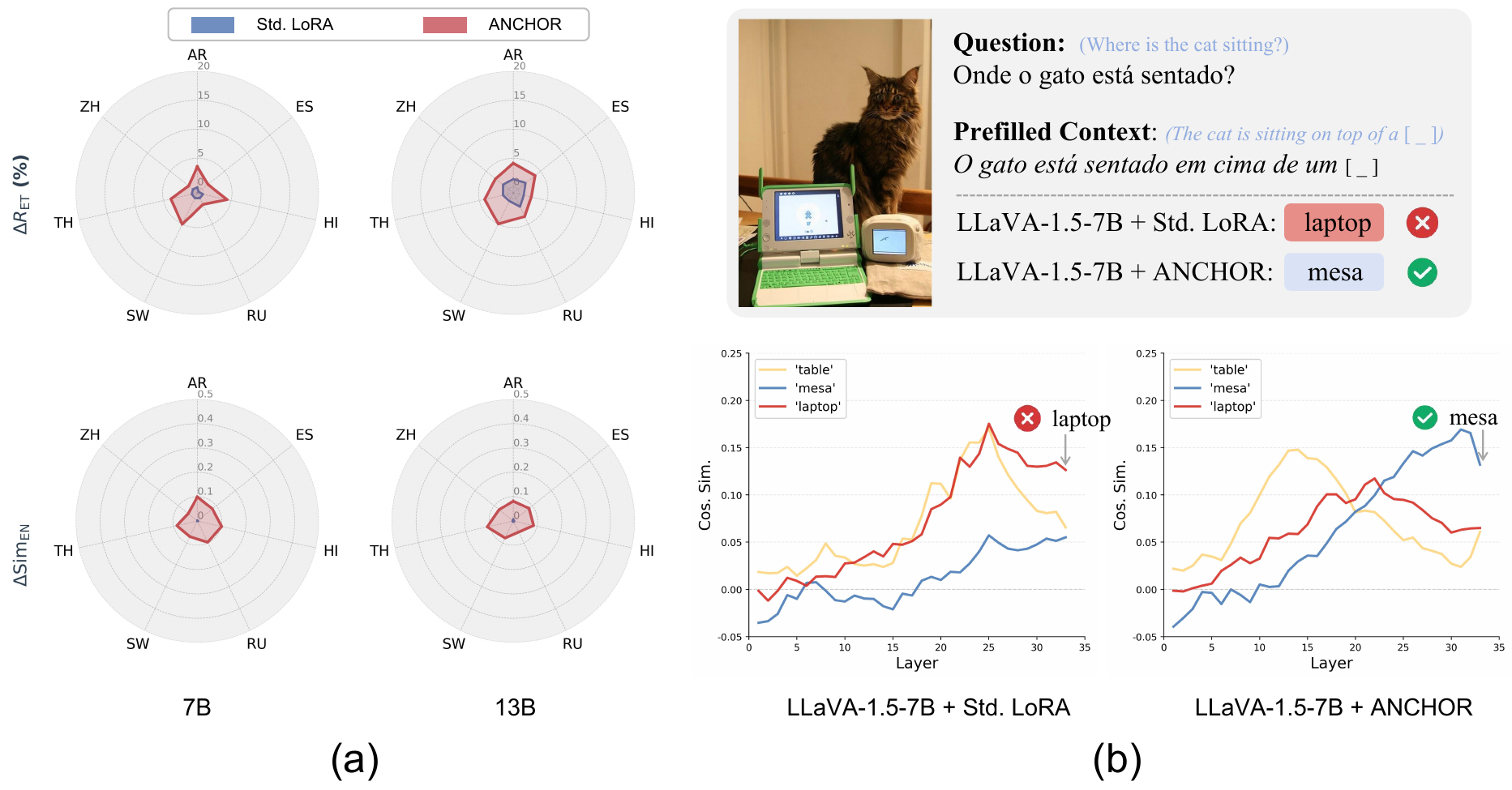}
    \caption{\textbf{(a)} Average visual causal effect ($\Delta R_{ET}$ and $\Delta \text{Sim}_{en}$) across 7 languages.\textbf{(b)} Layer-wise hidden state dynamics in a cross-lingual VQA case, showing how ANCHOR enables visual grounding to override textual priors during translation.}
    \label{fig:main_mechanism}
\end{figure}

\subsection{Mechanistic Analysis: Validating Ghost Anchor Rectification}
\label{sec:quantifying_visual_influence}

We mechanistically validate how the ANCHOR framework rectifies the Ghost Anchor phenomenon through (1) quantifying the visual causal influence on the early translation stage, and (2) examining a layer-wise case study where visual anchoring successfully overrides incorrect linguistic priors.

\paragraph{Enabling Visual Causal Effect} Following the methodology in \cref{sec:causal_validation}, we measure the visual causal effect ($\Delta R_{ET}$ and $\Delta Sim_{en}$) across the early layers ($l \in [1, 10]$). As shown in \cref{fig:main_mechanism}(a), standard LLaVA-1.5 exhibits limited visual influence, with baseline $\Delta R_{ET}$ averaging near zero ($-0.016\%$ for 7B and $1.18\%$ for 13B). This provides evidence consistent with the Ghost Anchor effect. In contrast, ANCHOR produces a positive shift in both $\Delta R_{ET}$ and $\Delta \text{Sim}_{en}$ across model scales, indicating increased visual influence on the measured early translation process.

\paragraph{Case Study: Visual Anchoring Overriding Linguistic Priors} To contextualize this shift, we examine a cross-lingual VQA case where the baseline follows a text-induced prior and outputs \textit{laptop} rather than the visual target, a \textit{table} (\textit{mesa} in Portuguese). To trace this internal semantic evolution, we compute the layer-wise cosine similarity between the final prompt token's normalized hidden state and the static unembedding weight vectors of specific target concepts. As shown in \cref{fig:main_mechanism}(b), the baseline's early-layer hidden states exhibit stronger alignment with the misleading linguistic prior. In contrast, ANCHOR shifts the internal trajectory toward the visually supported English concept (\textit{table}), which is consistent with its final Portuguese output (\textit{mesa}).

\section{Limitations}
Our analysis operationalizes the English Semantic Manifold as an empirical property of the English-dominant MLLMs studied in this work, rather than as a universal or normative semantic space. Models trained with different language distributions or architectural designs may exhibit different intermediate semantic geometries. In addition, our mechanistic metrics, including the logit lens, English Translation Ratio, and Visual Grounding Score, measure the decodability of semantic information from intermediate representations, which does not necessarily imply that the model uses all decoded information during generation. Although the Gaussian-noise intervention provides controlled evidence about the influence of visual content, it also introduces a distribution shift and may alter input properties beyond high-level semantics. Therefore, our results characterize modality asynchrony under these operationalizations rather than providing an exhaustive account of all cross-modal causal pathways.

\section{Conclusion}
In this work, we investigate why visual signals provide limited assistance as cross-lingual bridges for multilingual MLLMs. Mechanistic analysis identifies the Ghost Anchor phenomenon associated with Modality Asynchrony: linguistic streams quickly converge to an English-centric semantic manifold in early layers, whereas visual semantics emerge later, limiting the influence of visual inputs during early cross-modal alignment. To address this, we propose ANCHOR, which leverages Proactive Visual Anchoring (PVA) with an external Visual Foundation Model to accelerate early-layer visual semantic emergence and encourage visual evidence to guide linguistic translation. Experiments on xMMMU, MaXM, and CVQA show that ANCHOR improves cross-lingual visual reasoning and mitigates the performance degradation associated with unanchored early-layer adaptation.

\section*{Source Availability}
The datasets and code are available at \url{https://github.com/YihDu/MM_align}.

\section*{Acknowledgements}
This work is supported by Longgang District Special Funds for Science and Technology Innovation under Grant LGXXKYJG2024001.

\newpage
\bibliography{colm2026_conference}

@inproceedings{LRM-LLaVA,
  title={LRM-LLaVA: Overcoming the Modality Gap of Multilingual Large Language-Vision Model for Low-Resource Languages},
  author={Li, Junchen and Yang, Qing and Jiang, Bojian and Zhu, Shaolin and Sun, Qingxuan},
  booktitle={Proceedings of the AAAI Conference on Artificial Intelligence},
  pages={24449--24457},
  year={2025}
}

@article{AyaVision,
  title={Aya Vision: Advancing the Frontier of Multilingual Multimodality},
  author={Dash, Saurabh and Nan, Yiyang and Dang, John and Ahmadian, Arash and Singh, Shivalika and Smith, Madeline and Venkitesh, Bharat and Shmyhlo, Vlad and Aryabumi, Viraat and Beller-Morales, Walter and others},
  journal={arXiv preprint arXiv:2505.08751},
  year={2025}
}

@inproceedings{Pangea,
  title={Pangea: A fully open multilingual multimodal llm for 39 languages},
  author={Yue, Xiang and Song, Yueqi and Asai, Akari and Kim, Seungone and de Dieu Nyandwi, Jean and Khanuja, Simran and Kantharuban, Anjali and Sutawika, Lintang and Ramamoorthy, Sathyanarayanan and Neubig, Graham},
  booktitle={The Thirteenth International Conference on Learning Representations},
  year={2024}
}

@article{semantic-hub,
  title={The semantic hub hypothesis: Language models share semantic representations across languages and modalities},
  author={Wu, Zhaofeng and Yu, Xinyan Velocity and Yogatama, Dani and Lu, Jiasen and Kim, Yoon},
  journal={arXiv preprint arXiv:2411.04986},
  year={2024}
}

@inproceedings{M3P,
  title={M3p: Learning universal representations via multitask multilingual multimodal pre-training},
  author={Ni, Minheng and Huang, Haoyang and Su, Lin and Cui, Edward and Bharti, Taroon and Wang, Lijuan and Zhang, Dongdong and Duan, Nan},
  booktitle={Proceedings of the IEEE/CVF conference on computer vision and pattern recognition},
  pages={3977--3986},
  year={2021}
}

@inproceedings{Uc2,
  title={Uc2: Universal cross-lingual cross-modal vision-and-language pre-training},
  author={Zhou, Mingyang and Zhou, Luowei and Wang, Shuohang and Cheng, Yu and Li, Linjie and Yu, Zhou and Liu, Jingjing},
  booktitle={Proceedings of the IEEE/CVF Conference on Computer Vision and Pattern Recognition},
  pages={4155--4165},
  year={2021}
}

@inproceedings{huang2021multilingual,
  title={Multilingual multimodal pre-training for zero-shot cross-lingual transfer of vision-language models},
  author={Huang, Po-Yao and Patrick, Mandela and Hu, Junjie and Neubig, Graham and Metze, Florian and Hauptmann, Alexander G},
  booktitle={Proceedings of the 2021 Conference of the North American Chapter of the Association for Computational Linguistics: Human Language Technologies},
  pages={2443--2459},
  year={2021}
}

@article{sundar2025steering,
  title={Steering into new embedding spaces: Analyzing cross-lingual alignment induced by model interventions in multilingual language models},
  author={Sundar, Anirudh and Williamson, Sinead and Metcalf, Katherine and Theobald, Barry-John and Seto, Skyler and Fedzechkina, Masha},
  journal={arXiv preprint arXiv:2502.15639},
  year={2025}
}

@inproceedings{xGQA,
  title={xGQA: Cross-lingual visual question answering},
  author={Pfeiffer, Jonas and Geigle, Gregor and Kamath, Aishwarya and Steitz, Jan-Martin O and Roth, Stefan and Vuli{\'c}, Ivan and Gurevych, Iryna},
  booktitle={Findings of the association for computational linguistics: ACL 2022},
  pages={2497--2511},
  year={2022}
}

@inproceedings{Dataset_COCO,
  title={Microsoft coco: Common objects in context},
  author={Lin, Tsung-Yi and Maire, Michael and Belongie, Serge and Hays, James and Perona, Pietro and Ramanan, Deva and Doll{\'a}r, Piotr and Zitnick, C Lawrence},
  booktitle={European conference on computer vision},
  pages={740--755},
  year={2014},
  organization={Springer}
}

@article{schut2025multilingual,
  title={Do Multilingual LLMs Think In English?},
  author={Schut, Lisa and Gal, Yarin and Farquhar, Sebastian},
  journal={arXiv preprint arXiv:2502.15603},
  year={2025}
}

@article{zhao2024large,
  title={How do large language models handle multilingualism?},
  author={Zhao, Yiran and Zhang, Wenxuan and Chen, Guizhen and Kawaguchi, Kenji and Bing, Lidong},
  journal={Advances in Neural Information Processing Systems},
  volume={37},
  pages={15296--15319},
  year={2024}
}

@article{venhoff2025visual,
  title={How Visual Representations Map to Language Feature Space in Multimodal LLMs},
  author={Venhoff, Constantin and Khakzar, Ashkan and Joseph, Sonia and Torr, Philip and Nanda, Neel},
  journal={arXiv preprint arXiv:2506.11976},
  year={2025}
}

@article{bai2025qwen3,
  title={Qwen3-vl technical report},
  author={Bai, Shuai and Cai, Yuxuan and Chen, Ruizhe and Chen, Keqin and Chen, Xionghui and Cheng, Zesen and Deng, Lianghao and Ding, Wei and Gao, Chang and Ge, Chunjiang and others},
  journal={arXiv preprint arXiv:2511.21631},
  year={2025}
}

@misc{nostalgebraist2020interpreting,
  title={Interpreting GPT: The Logit Lens},
  author={Nostalgebraist},
  year={2020},
  month={Aug},
  howpublished={\url{https://www.alignmentforum.org/posts/AcKRB8wDpdaN6v6ru/interpreting-gpt-the-logit-lens}},
  note={Accessed: 5 Jan 2026} 
}

@inproceedings{chen2024sharegpt4v,
  title={Sharegpt4v: Improving large multi-modal models with better captions},
  author={Chen, Lin and Li, Jinsong and Dong, Xiaoyi and Zhang, Pan and He, Conghui and Wang, Jiaqi and Zhao, Feng and Lin, Dahua},
  booktitle={European Conference on Computer Vision},
  pages={370--387},
  year={2024},
  organization={Springer}
}

@article{romero2024cvqa,
  title={Cvqa: Culturally-diverse multilingual visual question answering benchmark},
  author={Romero, David and Lyu, Chenyang and Wibowo, Haryo Akbarianto and Lynn, Teresa and Hamed, Injy and Kishore, Aditya Nanda and Mandal, Aishik and Dragonetti, Alina and Abzaliev, Artem and Tonja, Atnafu Lambebo and others},
  journal={arXiv preprint arXiv:2406.05967},
  year={2024}
}

@inproceedings{changpinyo2023maxm,
  title={Maxm: Towards multilingual visual question answering},
  author={Changpinyo, Soravit and Xue, Linting and Yarom, Michal and Thapliyal, Ashish and Szpektor, Idan and Amelot, Julien and Chen, Xi and Soricut, Radu},
  booktitle={Findings of the Association for Computational Linguistics: EMNLP 2023},
  pages={2667--2682},
  year={2023}
}

@article{GPT5ModelCard,
  title={Openai gpt-5 system card},
  author={Singh, Aaditya and Fry, Adam and Perelman, Adam and Tart, Adam and Ganesh, Adi and El-Kishky, Ahmed and McLaughlin, Aidan and Low, Aiden and Ostrow, AJ and Ananthram, Akhila and others},
  journal={arXiv preprint arXiv:2601.03267},
  year={2025}
}

@article{tong2026beyond,
  title={Beyond Language Modeling: An Exploration of Multimodal Pretraining},
  author={Tong, Shengbang and Fan, David and Nguyen, John and Brown, Ellis and Zhou, Gaoyue and Qian, Shengyi and Zheng, Boyang and Vallaeys, Th{\'e}ophane and Han, Junlin and Fergus, Rob and others},
  journal={arXiv preprint arXiv:2603.03276},
  year={2026}
}

@article{shen2025vl,
  title={VL-SAE: Interpreting and Enhancing Vision-Language Alignment with a Unified Concept Set},
  author={Shen, Shufan and Sun, Junshu and Huang, Qingming and Wang, Shuhui},
  journal={arXiv preprint arXiv:2510.21323},
  year={2025}
}

@article{shu2025large,
  title={Large vision-language model alignment and misalignment: A survey through the lens of explainability},
  author={Shu, Dong and Zhao, Haiyan and Hu, Jingyu and Liu, Weiru and Payani, Ali and Cheng, Lu and Du, Mengnan},
  journal={arXiv preprint arXiv:2501.01346},
  year={2025}
}

@article{zhong2024beyond,
  title={Beyond english-centric llms: What language do multilingual language models think in?},
  author={Zhong, Chengzhi and Cheng, Fei and Liu, Qianying and Jiang, Junfeng and Wan, Zhen and Chu, Chenhui and Murawaki, Yugo and Kurohashi, Sadao},
  journal={arXiv preprint arXiv:2408.10811},
  year={2024}
}

@article{song2025bridge,
  title={How to bridge the gap between modalities: Survey on multimodal large language model},
  author={Song, Shezheng and Li, Xiaopeng and Li, Shasha and Zhao, Shan and Yu, Jie and Ma, Jun and Mao, Xiaoguang and Zhang, Weimin and Wang, Meng},
  journal={IEEE Transactions on Knowledge and Data Engineering},
  volume={37},
  number={9},
  pages={5311--5329},
  year={2025},
  publisher={IEEE}
}

@article{Qwen2.5-VL,
  title={Qwen2.5-VL Technical Report},
  author={Bai, Shuai and Chen, Keqin and Liu, Xuejing and Wang, Jialin and Ge, Wenbin and Song, Sibo and Dang, Kai and Wang, Peng and Wang, Shijie and Tang, Jun and Zhong, Humen and Zhu, Yuanzhi and Yang, Mingkun and Li, Zhaohai and Wan, Jianqiang and Wang, Pengfei and Ding, Wei and Fu, Zheren and Xu, Yiheng and Ye, Jiabo and Zhang, Xi and Xie, Tianbao and Cheng, Zesen and Zhang, Hang and Yang, Zhibo and Xu, Haiyang and Lin, Junyang},
  journal={arXiv preprint arXiv:2502.13923},
  year={2025}
}

@misc{liu2024llavanext,
    title={LLaVA-NeXT: Improved reasoning, OCR, and world knowledge},
    url={https://llava-vl.github.io/blog/2024-01-30-llava-next/},
    author={Liu, Haotian and Li, Chunyuan and Li, Yuheng and Li, Bo and Zhang, Yuanhan and Shen, Sheng and Lee, Yong Jae},
    month={January},
    year={2024}
}

@misc{gemmateam2025gemma3technicalreport,
      title={Gemma 3 Technical Report}, 
      author={{Gemma Team}},
      year={2025},
      eprint={2503.19786},
      archivePrefix={arXiv},
      primaryClass={cs.CL},
      url={https://arxiv.org/abs/2503.19786}, 
}

@inproceedings{bugliarello2022iglue,
 title={IGLUE: A benchmark for transfer learning across modalities, tasks, and languages},
  author={Bugliarello, Emanuele and Liu, Fangyu and Pfeiffer, Jonas and Reddy, Siva and Elliott, Desmond and Ponti, Edoardo Maria and Vuli{\'c}, Ivan},
  booktitle={International Conference on Machine Learning},
  pages={2370--2392},
  year={2022},
  organization={PMLR}
}

@article{zheng2025mma,
  title={MMA-ASIA: A Multilingual and Multimodal Alignment Framework for Culturally-Grounded Evaluation},
  author={Zheng, Weihua and Liu, Zhengyuan and Chakraborty, Tanmoy and Xu, Weiwen and Gao, Xiaoxue and Tan, Bryan Chen Zhengyu and Zou, Bowei and Liu, Chang and Hu, Yujia and Xie, Xing and others},
  journal={arXiv preprint arXiv:2510.08608},
  year={2025}
}

@inproceedings{zeng2025converging,
  title={Converging to a lingua franca: Evolution of linguistic regions and semantics alignment in multilingual large language models},
  author={Zeng, Hongchuan and Han, Senyu and Chen, Lu and Yu, Kai},
  booktitle={Proceedings of the 31st International Conference on Computational Linguistics},
  pages={10602--10617},
  year={2025}
}

@article{elshabrawy2025alignment,
  title={When Alignment Hurts: Decoupling Representational Spaces in Multilingual Models},
  author={Elshabrawy, Ahmed and Kaing, Hour and Song, Haiyue and Aji, Alham Fikri and Tanaka, Hideki and Utiyama, Masao and Dabre, Raj},
  journal={arXiv preprint arXiv:2508.12803},
  year={2025}
}

@inproceedings{wendler-etal-2024-llamas,
    title = "Do Llamas Work in {E}nglish? On the Latent Language of Multilingual Transformers",
    author = "Wendler, Chris  and
      Veselovsky, Veniamin  and
      Monea, Giovanni  and
      West, Robert",
    editor = "Ku, Lun-Wei  and
      Martins, Andre  and
      Srikumar, Vivek",
    booktitle = "Proceedings of the 62nd Annual Meeting of the Association for Computational Linguistics (Volume 1: Long Papers)",
    month = aug,
    year = "2024",
    address = "Bangkok, Thailand",
    publisher = "Association for Computational Linguistics",
    url = "https://aclanthology.org/2024.acl-long.820/",
    doi = "10.18653/v1/2024.acl-long.820",
    pages = "15366--15394"
}

@inproceedings{geigle-etal-2024-mblip,
    title = "m{BLIP}: Efficient Bootstrapping of Multilingual Vision-{LLM}s",
    author = "Geigle, Gregor  and
      Jain, Abhay  and
      Timofte, Radu  and
      Glava{\v{s}}, Goran",
    editor = "Gu, Jing  and
      Fu, Tsu-Jui (Ray)  and
      Hudson, Drew  and
      Celikyilmaz, Asli  and
      Wang, William",
    booktitle = "Proceedings of the 3rd Workshop on Advances in Language and Vision Research (ALVR)",
    month = aug,
    year = "2024",
    address = "Bangkok, Thailand",
    publisher = "Association for Computational Linguistics",
    url = "https://aclanthology.org/2024.alvr-1.2/",
    doi = "10.18653/v1/2024.alvr-1.2",
    pages = "7--25"
}

@inproceedings{10.1145/3696410.3714861,
author = {Ahmat, Ahtamjan and Wang, Lei and Yang, Yating and Ma, Bo and Dong, Rui and Lu, Kaiwen and Ma, Rong and Wang, Xinyue},
title = {M2-VLP: Enhancing Multilingual Vision-Language Pre-Training via Multi-Grained Alignment},
year = {2025},
isbn = {9798400712746},
publisher = {Association for Computing Machinery},
address = {New York, NY, USA},
url = {https://doi.org/10.1145/3696410.3714861},
doi = {10.1145/3696410.3714861},
booktitle = {Proceedings of the ACM on Web Conference 2025},
pages = {3438–3450},
numpages = {13},
location = {Sydney NSW, Australia},
series = {WWW '25}
}

@article{tschannen2025siglip,
  title={Siglip 2: Multilingual vision-language encoders with improved semantic understanding, localization, and dense features},
  author={Tschannen, Michael and Gritsenko, Alexey and Wang, Xiao and Naeem, Muhammad Ferjad and Alabdulmohsin, Ibrahim and Parthasarathy, Nikhil and Evans, Talfan and Beyer, Lucas and Xia, Ye and Mustafa, Basil and others},
  journal={arXiv preprint arXiv:2502.14786},
  year={2025}
}
\bibliographystyle{colm2026_conference}

\newpage
\appendix

\section{Experimental Setup for Mechanistic Analysis}
\label{sec:exp_setup}

\subsection{Input Data Construction}
We randomly sample 100 images from the COCO 2017 Validation Set~\cite{Dataset_COCO} as our experimental stimuli. To ensure strict semantic consistency across languages, we develop a high-quality parallel corpus by translating all ground-truth English captions into seven diverse target languages (\cref{tab:languages}) using GPT-5.1~\cite{GPT5ModelCard}.

\begin{table}[htbp]
\centering
\setlength{\tabcolsep}{18pt}
\begin{tabular}{lc}
\toprule
\textbf{Language} & \textbf{Abbreviations} \\ \midrule
Chinese  & zh \\
Spanish  & es \\
Hindi    & hi \\
Arabic   & ar \\
Russian  & ru \\
Swahili  & sw \\
Thai     & th \\ \bottomrule
\end{tabular}
\caption{Selected languages used in the mechanistic experiments and their corresponding ISO 639-1 codes.}
\label{tab:languages}
\end{table}

\begin{figure*}
    \centering
    \includegraphics[width=\linewidth]{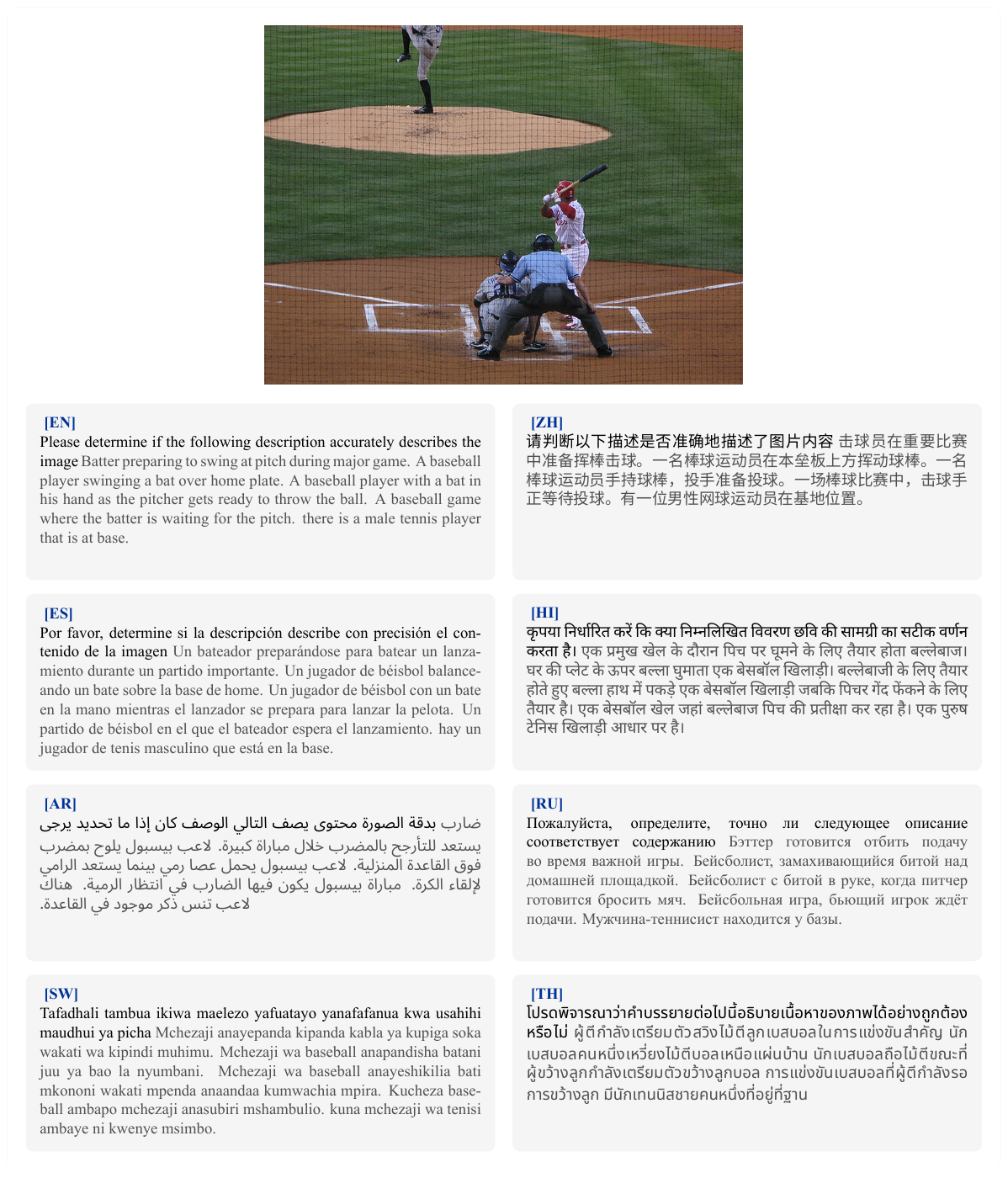}
    \caption{\textbf{Illustrative example of the input construction.} This figure demonstrates the parallel input constructions across 7 structurally diverse target languages. To enforce a strict monolingual context and prevent English prior leakage during inference, both the standardized verification instruction and the ground-truth image caption are translated into the target language and paired with identical visual tokens from the COCO dataset.}
    \label{fig:input_example_coco}
\end{figure*}

To maintain a pure monolingual context and prevent any English prior leakage, the input sequence is formulated entirely in the target language: \texttt{[Visual Tokens] + [Target Language Instruction] + [Target Language Caption]}. The instruction prompts the model to verify semantic consistency (e.g., \textit{"Please determine if the following description accurately describes the image."}, translated accordingly). An illustrative example of a constructed input sample across multiple languages is presented in \cref{fig:input_example_coco}.

\subsection{Hidden State Extraction}
To measure the linguistic and visual streams separately, we extract hidden states at each layer $l$ and isolate two subsets from the full sequence:

\begin{itemize}[leftmargin=1em]
    \item \textbf{Text tokens:} All hidden states corresponding to the multilingual instruction and caption.
    \item \textbf{Visual tokens:} Only hidden states from image patches that contain labeled foreground objects (identified via COCO segmentation masks), excluding ambiguous background regions.
\end{itemize}

Formally, we define the source-language target set as $\mathcal{H}_{text}^{(l)} = \{\mathbf{h}_{src,1,l}, \dots, \mathbf{h}_{src,M_{src},l}\}$, containing all $M_{src}$ source-language instruction and caption tokens. For English Similarity, the structurally matched English reference set is $\mathcal{H}_{en}^{(l)} = \{\mathbf{h}_{en,1,l}, \dots, \mathbf{h}_{en,M_{en},l}\}$, containing the corresponding $M_{en}$ English instruction and caption tokens. The visual target set is $\mathcal{H}_{vis}^{(l)} = \{\mathbf{h}_{vis,i,l} \mid i \in \mathcal{N}_{fg}\}$, where $\mathcal{N}_{fg}$ denotes the index set of labeled object tokens. These subsets are directly utilized to compute the metrics defined in Section~\ref{sec:metrics}.

\subsection{Metrics Overview}
\label{sec:metrics}
We quantify layer-wise dynamics through three complementary metrics: English Similarity ($\text{Sim}_{en}$) and English Translation Ratio ($R_{ET}$) for the text stream, and Visual Grounding Score ($S_{VG}$) for the vision stream. For vocabulary-level projections, we employ the \textit{logit lens} technique~\cite{nostalgebraist2020interpreting}, which maps an intermediate hidden state $\mathbf{h}_l$ to the probability distribution over the vocabulary $\mathcal{V}$ via the pre-trained unembedding matrix $\mathbf{W}_U \in \mathbb{R}^{|\mathcal{V}| \times d}$.

\subsubsection{English Similarity ($\text{Sim}_{en}$)}
To measure continuous geometric alignment toward the English semantic manifold $\mathcal{S}_{en}$, we first compute the sequence-level mean representation for the source-language input (using $\mathcal{H}_{text}^{(l)}$) and its structurally matched English reference (using $\mathcal{H}_{en}^{(l)}$). Both inputs contain the corresponding instruction and caption, while their token counts may differ across languages:
\begin{equation}
\overline{\mathbf{h}}_{src,l} = \frac{1}{M_{src}}\sum_{j=1}^{M_{src}} \mathbf{h}_{src,j,l}, \quad \overline{\mathbf{h}}_{en,l} = \frac{1}{M_{en}}\sum_{j=1}^{M_{en}} \mathbf{h}_{en,j,l}
\end{equation}
The layer-wise English Similarity is then defined as their cosine similarity:
\begin{equation}
    \text{Sim}_{en}(l) = \frac{\overline{\mathbf{h}}_{src,l} \cdot \overline{\mathbf{h}}_{en,l}}{\|\overline{\mathbf{h}}_{src,l}\| \|\overline{\mathbf{h}}_{en,l}\|}
\end{equation}
A low $\text{Sim}_{en}$ in initial layers reflects the geometric separation between source language manifolds and the English semantic pivot.

\subsubsection{English Translation Ratio ($R_{ET}$)}
To quantify vocabulary-level convergence, we analyze the proportion of text tokens in $\mathcal{H}_{text}^{(l)}$ that decode into English. For the $j$-th source language token, we extract the top-$k$ candidate decoded words via the logit lens to form the set $\mathcal{T}_{j,l}$, setting $k=5$:
\begin{equation}
\mathcal{T}_{j,l} = \text{Top}_{k}(\text{Softmax}(\mathbf{W}_U \mathbf{h}_{src,j,l}))
\end{equation}
To rigorously exclude language-agnostic symbols (e.g., punctuation, numbers), we employ GPT-4o as a zero-shot language classifier, denoted as $\Phi_{lang}(\cdot)$. A token is considered to have transitioned into the English space if at least one of its top-$k$ decoded words is classified as English. The translation ratio is calculated only over tokens that bear actual linguistic content---that is, tokens for which at least one top-$k$ decoded word is classified as a natural language (e.g., English, Chinese, Arabic) rather than as punctuation, numbers, or other symbols. Formally:
\begin{equation}
R_{ET}(l) = \frac{\sum_{j=1}^{M_{src}} \mathbb{I} \Big[ \exists w \in \mathcal{T}_{j,l} \text{ s.t. } \Phi_{lang}(w) = \text{English} \Big]}{\sum_{j=1}^{M_{src}} \mathbb{I} \Big[ \exists w \in \mathcal{T}_{j,l} \text{ s.t. } \Phi_{lang}(w) \in \{\text{Natural Languages}\} \Big]}
\end{equation}
This metric directly quantifies the discrete lexical transition into the English space while accounting for the distributional nature of the output logits.

\subsubsection{Visual Grounding Score ($S_{VG}$)}
We evaluate the emergence of visual semantics by measuring how visual tokens align with concrete linguistic concepts. Using the labeled object hidden states $\mathcal{H}_{vis}^{(l)}$ extracted via COCO annotations, we monitor their convergence toward the set of ground-truth object labels $\mathcal{C}_{GT}$.

For each object token $i \in \mathcal{N}_{fg}$ at layer $l$, we extract the top-$k$ candidate text tokens via the logit lens: $\mathcal{T}_{i,l} = \text{Top}_{k}(\text{Softmax}(\mathbf{W}_{U} \mathbf{h}_{vis,i,l}))$, setting $k=5$. The grounding score $S_{VG}$ is defined as the successful semantic hit rate across these object tokens:
\begin{equation}
S_{VG}(l) = \frac{1}{|\mathcal{N}_{fg}|} \sum_{i \in \mathcal{N}_{fg}} \mathbb{I} \Big[ \exists w \in \mathcal{T}_{i,l}, \exists c \in \mathcal{C}_{GT} \text{ s.t. } w \subseteq c \Big]
\end{equation}
where $w \subseteq c$ denotes that a decoded candidate word matches or is a valid substring of a ground-truth object label. $S_{VG}$ serves as a direct measurement for the cross-modal semantic verbalization process.

\subsection{Noise Intervention: Implementation Details}
\label{sup:noise_intervention}

\textbf{Motivation.} The noise intervention creates a control condition in which visual tokens remain present while recognizable high-level visual content is substantially reduced. This probes the sensitivity of linguistic translation to visual content while holding constant: (i) sequence length, (ii) token positions, and (iii) per-channel first- and second-order pixel statistics.

\textbf{Implementation.} For each image $I$, we compute per-channel mean $\mu_c$ and standard deviation $\sigma_c$ from the original pixel values. The noise image $I_{\text{noise}}$ is generated by sampling from a Gaussian distribution with matching statistics:
\begin{equation}
I_{\text{noise}}^{(c)}(x,y) \sim \mathcal{N}\left(\mu_c, \sigma_c^2\right), \quad \text{where } \mu_c = \frac{1}{HW}\sum_{x,y} I^{(c)}(x,y)
\end{equation}
The noise tensor is passed to the vision encoder via \texttt{pixel\_values}, preserving the input dimensions, patch count, and visual-token positions while disrupting recognizable object and scene structure.

\begin{figure*}
    \centering
    \includegraphics[width=\linewidth]{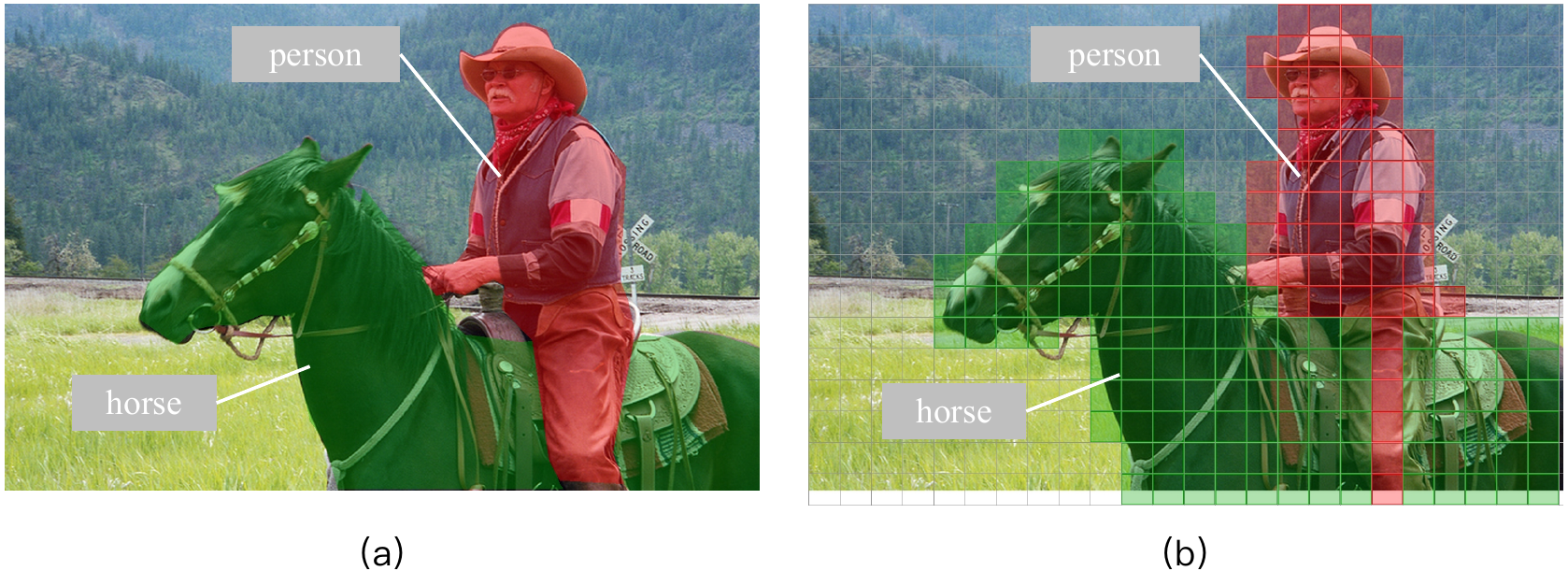}
    \caption{\textbf{(a)} Original segmentation masks and semantic labels from the COCO dataset. \textbf{(b)} The mapped $24 \times 24$ visual patches and their corresponding labels.}
    \label{fig:mask_example}
\end{figure*}

\begin{figure*}[h]
    \centering
    \includegraphics[width=0.6\linewidth]{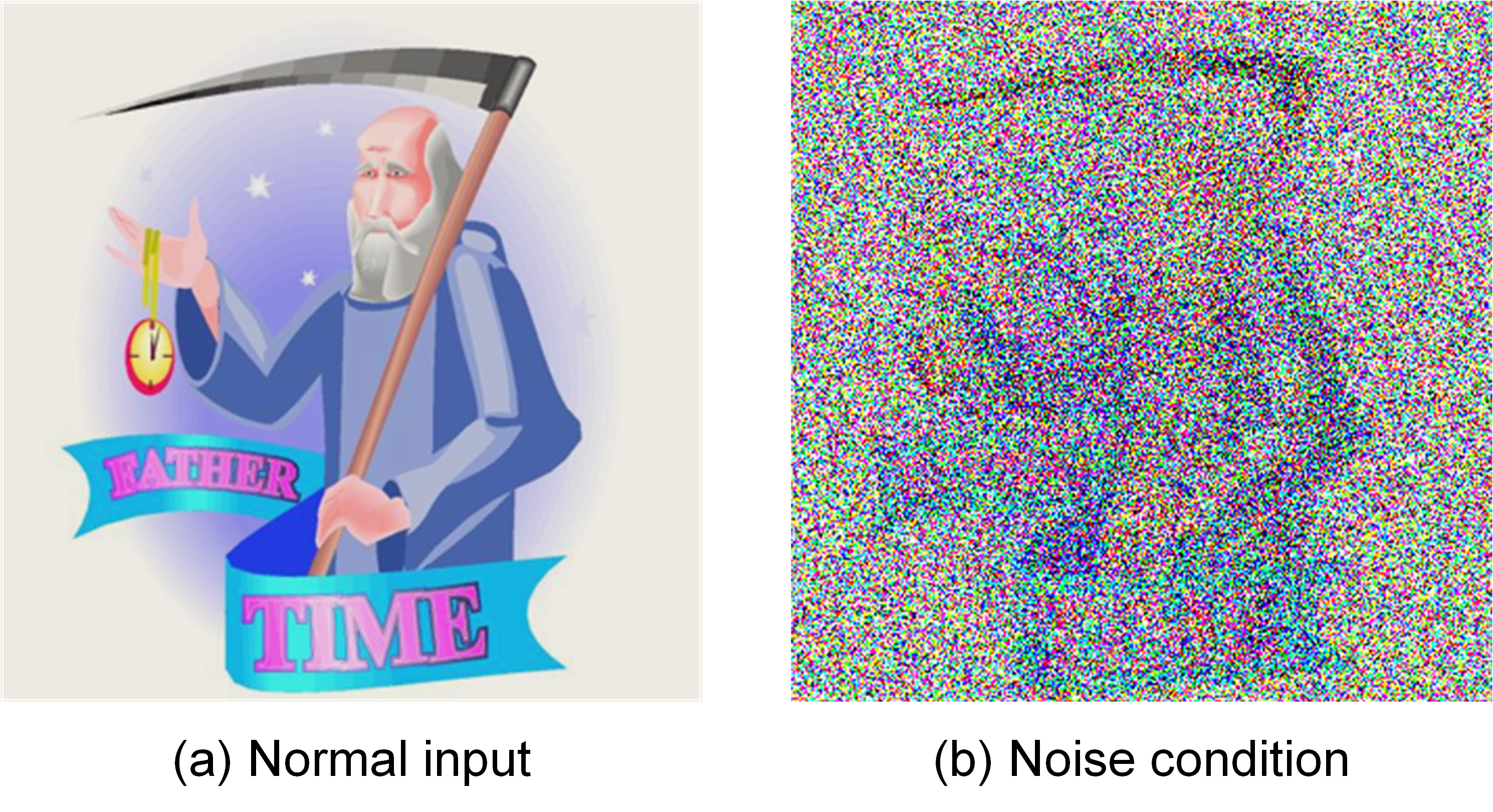}
    \caption{Visual comparison of intervention conditions. \textbf{Left:} Normal RGB input with intact semantic content. \textbf{Right:} Gaussian noise with matched per-channel statistics ($\mu$, $\sigma$) but no recognizable objects or scenes.}
    \label{fig:Noise_condition_example}
\end{figure*}

\textbf{Validation.} As shown in \cref{fig:Noise_condition_example}, the noise intervention removes recognizable object contours, textures, and scene structure from the input. Because the matched noise preserves only per-channel first- and second-order pixel statistics, it provides a control condition with substantially reduced high-level visual semantics; we do not assume that it eliminates every decodable visual signal.

\section{Datasets}
\label{sec:dataset}

\subsection{Training Datasets}
\label{sec:train_dataset}
To construct a high-quality multilingual visual instruction tuning dataset, we systematically sample and filter image-text pairs from the ShareGPT4V dataset~\cite{chen2024sharegpt4v}. To minimize potential noise and artifacts introduced during the machine translation process, we implement a strict filtering mechanism based on text length and annotation quality, prioritizing structurally straightforward sentences. Specifically, we extract 9,000 image-text pairs from the COCO subset and 6,000 from the GQA subset within ShareGPT4V. These specific subsets are chosen because their concise and visually-grounded question-answering formats make them ideal anchors for cross-lingual alignment. Subsequently, we translate the selected English questions and answers into nine typologically diverse target languages using GPT-5.1 \citep{GPT5ModelCard}. These languages, namely Arabic (ar), Spanish (es), French (fr), Hindi (hi), Portuguese (pt), Russian (ru), Swahili (sw), Thai (th), and Chinese (zh), are specifically chosen to encompass a broad spectrum of language families and resource availability levels. This pipeline ensures that the semantic consistency between the visual inputs and text representations is accurately preserved across different linguistic contexts.

\subsection{Details of Evaluated Multilingual Multimodal Benchmarks}
\label{sec:benchmark_details}

To comprehensively assess visual understanding and reasoning capabilities across diverse linguistic and cultural contexts, we evaluate the proposed method on three distinct multilingual multimodal benchmarks: xMMMU \cite{Pangea}, MaXM \cite{changpinyo2023maxm}, and CVQA \cite{romero2024cvqa}.

\paragraph{Multilingual MMMU (xMMMU)} 
xMMMU \cite{Pangea} focuses on complex multimodal understanding and reasoning across multiple academic subjects, evaluating the capacity to understand specialized content across different languages and modalities. The dataset features 183 subfields and 30 diverse image types, including charts, diagrams, and chemical structures. It consists of 300 English questions randomly sampled from the MMMU validation set, which were subsequently translated into six languages (ar, fr, hi, id, ja, pt) using GPT-4o. The questions are presented in multiple-choice and short-answer formats. We follow the evaluation protocol in \cite{Pangea} and report accuracy.

\paragraph{Multilingual Visual Question Answering (MaXM)} 
MaXM \cite{changpinyo2023maxm} is a test-only multilingual open-ended visual question-answering benchmark encompassing seven languages (fr, hi, th, zh, en, he, ro). To address the challenge of cultural diversity in multimodal understanding, the images used in MaXM were taken in regions where each language is spoken to better reflect cultural contexts. It contains 2K questions, where answers are in the same language as the question. We use accuracy as the evaluation metric following prior work \cite{Pangea, AyaVision}.

\paragraph{Culturally-diverse Multilingual Visual Question Answering (CVQA)}
CVQA \cite{romero2024cvqa} evaluates the ability to reason about culturally diverse visual content. It consists of over 21K multiple-choice questions designed to test multimodal reasoning across 39 distinct language-region pairs, covering 31 unique languages and 30 countries. To accurately reflect global cultural nuances, the dataset incorporates a wide spectrum of languages, ranging from high-resource languages like Chinese and regional variations of Spanish, to lower-resource indigenous languages such as Amharic, Javanese, Minangkabau, and Swahili. We evaluate models under the local-language question setting and report accuracy as the evaluation metric. 

\section{Additional Experiment Results}
\subsection{Mechanistic Results on More MLLMs}
\label{sec:appendix_sota}
\Cref{tab:mechanistic_sota} reports the effect of visual intervention on the English Translation Ratio ($\Delta R_{\text{ET}}$) and the English Similarity ($\Delta \text{Sim}_{\text{en}}$), averaged over the early translation stage ($L \in [1, 10]$). Consistent with our main findings, all evaluated models, including Qwen2.5-VL~\cite{Qwen2.5-VL}, Gemma 3~\cite{gemmateam2025gemma3technicalreport}, and LLaVA-NeXT~\cite{liu2024llavanext}, exhibit small changes under the intervention, providing evidence that the Ghost Anchor phenomenon extends across the evaluated architectures and scales.

\begin{table}[ht]
\centering
\begin{tabular}{l cc}
\toprule
\textbf{Model} & \textbf{$\Delta R_{\text{ET}} \uparrow$} & \textbf{$\Delta \text{Sim}_{\text{en}} \uparrow$} \\
\midrule
Qwen2.5-VL 3B  & $4.12 \times 10^{-3}$   & $-2.31 \times 10^{-4}$ \\
Qwen2.5-VL 7B  & $5.34 \times 10^{-3}$   & $-4.51 \times 10^{-3}$ \\
Gemma-3-4B     & $-1.13 \times 10^{-2}$  & $-4.74 \times 10^{-4}$ \\
Gemma-3-13B    & $6.95 \times 10^{-3}$   & $7.37 \times 10^{-3}$ \\
LLaVA-NeXT-7B  & $-3.19 \times 10^{-3}$  & $-4.71 \times 10^{-4}$ \\
LLaVA-NeXT-13B & $8.61 \times 10^{-3}$   & $-8.28 \times 10^{-3}$ \\
\bottomrule
\end{tabular}
\caption{Visual Causal Effect on English Translation Ratio ($\Delta R_{\text{ET}}$) and English Similarity ($\Delta \text{Sim}_{\text{en}}$).}
\label{tab:mechanistic_sota}
\end{table}

\begin{table*}[t]
\centering
\small
\colorlet{trainedbg}{pink!30}
\colorlet{zerobg}{cyan!15}
\setlength{\tabcolsep}{11pt}
\setlength{\aboverulesep}{0pt}
\setlength{\belowrulesep}{0pt}
\renewcommand{\arraystretch}{1.2}
\begin{tabular}{lccccc}
\toprule
\textbf{Backbone} & \textbf{Size} & $\boldsymbol{\Delta}$\textbf{En}
& \cellcolor{trainedbg}$\boldsymbol{\Delta}$\textbf{SFT Avg. \faHotjar}
& \cellcolor{zerobg}$\boldsymbol{\Delta}$\textbf{Zero-shot Avg. \faSnowflake}
& $\boldsymbol{\Delta}$\textbf{Avg.} \\
\midrule
LLaVA-NeXT & 7B  & +0.4 & +3.1 & +1.7 & \textbf{+2.3} \\
LLaVA-NeXT & 13B & +0.5 & +4.0 & +1.9 & \textbf{+2.9} \\
\midrule
InternVL3  & 8B  & +0.2 & +2.3 & +1.1 & \textbf{+1.7} \\
InternVL3  & 14B & +0.5 & +2.6 & +1.3 & \textbf{+1.9} \\
\bottomrule
\end{tabular}
\caption{Generalization to newer MLLM backbones on xMMMU. Each entry reports the accuracy difference in percentage points between ANCHOR and the corresponding Std. LoRA baseline ($\text{ANCHOR}-\text{Std. LoRA}$). SFT Avg. and Zero-shot Avg. are macro-averages over their respective language groups, while Avg. is the macro-average over all seven evaluated language subsets, including English.}
\label{tab:newer_mllm_xmmmu}
\end{table*}

\begin{table*}[t]
\centering
\small
\colorlet{trainedbg}{pink!30}
\colorlet{zerobg}{cyan!15}
\setlength{\tabcolsep}{11pt}
\setlength{\aboverulesep}{0pt}
\setlength{\belowrulesep}{0pt}
\renewcommand{\arraystretch}{1.2}
\begin{tabular}{lccccc}
\toprule
\textbf{Backbone} & \textbf{Size} & $\boldsymbol{\Delta}$\textbf{En}
& \cellcolor{trainedbg}$\boldsymbol{\Delta}$\textbf{SFT Avg. \faHotjar}
& \cellcolor{zerobg}$\boldsymbol{\Delta}$\textbf{Zero-shot Avg. \faSnowflake}
& $\boldsymbol{\Delta}$\textbf{Avg.} \\
\midrule
LLaVA-NeXT & 7B  & +0.8 & +3.4 & +1.2 & \textbf{+2.4} \\
LLaVA-NeXT & 13B & +1.0 & +3.8 & +1.3 & \textbf{+2.7} \\
\midrule
InternVL3  & 8B  & +0.5 & +2.6 & +0.8 & \textbf{+1.8} \\
InternVL3  & 14B & +0.6 & +3.0 & +0.9 & \textbf{+2.1} \\
\bottomrule
\end{tabular}
\caption{Generalization to newer MLLM backbones on MaXM. Each entry reports the accuracy difference in percentage points between ANCHOR and the corresponding Std. LoRA baseline ($\text{ANCHOR}-\text{Std. LoRA}$). SFT Avg. and Zero-shot Avg. are macro-averages over their respective language groups, while Avg. is the macro-average over all seven evaluated language subsets, including English.}
\label{tab:newer_mllm_maxm}
\end{table*}

\begin{table*}[t]
\centering
\small
\colorlet{trainedbg}{pink!30}
\colorlet{zerobg}{cyan!15}
\setlength{\tabcolsep}{15pt}
\setlength{\aboverulesep}{0pt}
\setlength{\belowrulesep}{0pt}
\renewcommand{\arraystretch}{1.2}
\begin{tabular}{lcccc}
\toprule
\textbf{Backbone} & \textbf{Size}
& \cellcolor{trainedbg}$\boldsymbol{\Delta}$\textbf{SFT Avg. \faHotjar}
& \cellcolor{zerobg}$\boldsymbol{\Delta}$\textbf{Zero-shot Avg. \faSnowflake}
& $\boldsymbol{\Delta}$\textbf{Avg.} \\
\midrule
LLaVA-NeXT & 7B  & +3.4 & +1.4 & \textbf{+2.1} \\
LLaVA-NeXT & 13B & +3.8 & +1.8 & \textbf{+2.5} \\
\midrule
InternVL3  & 8B  & +2.6 & +1.0 & \textbf{+1.6} \\
InternVL3  & 14B & +2.9 & +1.1 & \textbf{+1.7} \\
\bottomrule
\end{tabular}
\caption{Generalization to newer MLLM backbones on CVQA. Each entry reports the accuracy difference in percentage points between ANCHOR and the corresponding Std. LoRA baseline ($\text{ANCHOR}-\text{Std. LoRA}$). SFT Avg. and Zero-shot Avg. are macro-averages over their respective language--region groups, while Avg. is the macro-average over all 39 language--region pairs.}
\label{tab:newer_mllm_cvqa}
\end{table*}

\clearpage

\subsection{Full Results on the CVQA Benchmark}
\colorlet{trainedbg}{pink!30} 
\colorlet{zerobg}{cyan!15} 
\begin{table}[h!]
\centering
\setlength{\aboverulesep}{0pt}
\setlength{\belowrulesep}{0pt}
\renewcommand{\arraystretch}{1.2}
\resizebox{\textwidth}{!}{
\begin{tabular}{l|cccccccccc}
\toprule
\textbf{Models} & \cellcolor{trainedbg}\textbf{ar-EG} & \cellcolor{trainedbg}\textbf{es-AR} & \cellcolor{trainedbg}\textbf{es-CL} & \cellcolor{trainedbg}\textbf{es-CO} & \cellcolor{trainedbg}\textbf{es-EC} & \cellcolor{trainedbg}\textbf{es-ES} & \cellcolor{trainedbg}\textbf{es-MX} & \cellcolor{trainedbg}\textbf{es-UY} & \cellcolor{trainedbg}\textbf{hi-IN} & \cellcolor{trainedbg}\textbf{pt-BR} \\
\midrule
LLaVA-1.5-7B & 32.5 & 54.3 & 60.7 & 56.4 & 53.6 & 67.0 & 48.3 & 43.5 & 40.3 & 57.4 \\
\quad + Std. LoRA & 31.5 & 51.7 & 56.4 & 55.2 & 49.5 & 62.0 & 46.1 & 40.6 & 38.8 & 56.3 \\
\quad + Early-Layer FFT & 29.6 & 41.5 & 44.9 & 46.5 & 42.3 & 50.3 & 43.7 & 37.8 & 37.3 & 45.8 \\
\rowcolor{gray!15} \textbf{\quad + ANCHOR} & 37.3 & 55.8 & 60.3 & 58.9 & 53.7 & 69.3 & 51.2 & 46.7 & 42.5 & 59.0 \\
\midrule
LLaVA-1.5-13B & 38.4 & 57.4 & 62.0 & 61.4 & 53.6 & 70.4 & 51.1 & 39.7 & 49.3 & 63.4 \\
\quad + Std. LoRA & 36.0 & 54.7 & 57.7 & 61.4 & 55.0 & 67.9 & 50.8 & 39.7 & 46.8 & 59.9 \\
\quad + Early-Layer FFT & 28.1 & 41.1 & 44.0 & 44.0 & 42.5 & 52.2 & 41.8 & 34.0 & 40.3 & 47.5 \\
\rowcolor{gray!15} \textbf{\quad + ANCHOR} & 40.1 & 59.2 & 65.3 & 63.0 & 56.1 & 70.8 & 51.5 & 42.8 & 50.3 & 64.3\\
\midrule
\textbf{Models} & \cellcolor{trainedbg}\textbf{ru-RU} & \cellcolor{trainedbg}\textbf{sw-KE} & \cellcolor{trainedbg}\textbf{zh-CN} & \cellcolor{trainedbg}\textbf{zh-SG} & \cellcolor{zerobg}\textbf{am-ET} & \cellcolor{zerobg}\textbf{bg-BG} & \cellcolor{zerobg}\textbf{bn-IN} & \cellcolor{zerobg}\textbf{br-FR} & \cellcolor{zerobg}\textbf{fil-PH} & \cellcolor{zerobg}\textbf{ga-IE} \\
\midrule
LLaVA-1.5-7B & 57.5 & 36.3 & 48.6 & 48.6 & 26.5 & 39.4 & 29.7 & 29.6 & 46.3 & 44.2 \\
\quad + Std. LoRA & 57.0 & 37.4 & 44.7 & 47.6 & 26.1 & 38.8 & 33.6 & 26.9 & 42.4 & 43.9 \\
\quad + Early-Layer FFT & 41.5 & 38.5 & 42.8 & 43.9 & 27.4 & 34.8 & 33.2 & 30.6 & 39.4 & 36.8 \\
\rowcolor{gray!15} \textbf{\quad + ANCHOR} & 57.1 & 36.3 & 49.1 & 52.5 & 26.4 & 38.0 & 34.6  & 28.3 & 44.5 & 44.0 \\
\midrule
LLaVA-1.5-13B & 57.0 & 42.9 & 54.3 & 52.8 & 25.2 & 44.7 & 34.6 & 35.8 & 48.8 & 46.6 \\
\quad + Std. LoRA & 53.5 & 41.8 & 50.5 & 52.8 & 26.1 & 42.3 & 32.9 & 32.1 & 47.8 & 42.0 \\
\quad + Early-Layer FFT &46.0 & 44.7 & 41.8 & 42.0 & 36.8 & 36.7 & 27.6 & 31.4 & 38.9 & 36.8 \\
\rowcolor{gray!15} \textbf{\quad + ANCHOR} & 57.4 & 46.0 & 59.4 & 56.0 & 26.0 & 42.8 & 35.7 & 34.7 & 49.1 & 48.5  \\
\midrule
\textbf{Models} & \cellcolor{zerobg}\textbf{id-ID} & \cellcolor{zerobg}\textbf{ig-NG} & \cellcolor{zerobg}\textbf{ja-JP} & \cellcolor{zerobg}\textbf{jv-ID} & \cellcolor{zerobg}\textbf{ko-KR} & \cellcolor{zerobg}\textbf{min-ID} & \cellcolor{zerobg}\textbf{mn-MN} & \cellcolor{zerobg}\textbf{mr-IN} & \cellcolor{zerobg}\textbf{ms-MY} & \cellcolor{zerobg}\textbf{no-NO} \\
\midrule
LLaVA-1.5-7B & 42.2 & 32.5 & 35.0 & 36.4 & 51.0 & 36.3 & 29.8 & 36.1 & 47.9 & 52.5 \\
\quad + Std. LoRA & 40.1 & 32.0 & 35.5 & 35.0 & 47.2 & 36.2 & 31.1 & 34.2 & 45.4 & 51.5 \\
\quad + Early-Layer FFT & 35.7 & 27.5 & 35.5 & 31.0 & 43.5 & 31.1 & 32.1 & 32.7 & 35.9 & 42.8 \\
\rowcolor{gray!15} \textbf{\quad + ANCHOR} & 40.4 & 31.9 & 36.9 & 37.3 & 49.3 & 37.1 & 33.2 & 37.1 & 47.2 & 53.1 \\
\midrule
LLaVA-1.5-13B & 46.6 & 36.5 & 45.8 & 36.7 & 54.1 & 35.9 & 30.8 & 38.6 & 48.6 & 58.2 \\
\quad + Std. LoRA & 40.8 & 31.5 & 42.4 & 37.7 & 48.6 & 36.7 & 31.0 & 39.1 & 44.8 & 52.2 \\
\quad + Early-Layer FFT &39.6 & 30.5 & 37.0 & 32.3 & 47.2 & 35.1 & 27.2 & 35.2 & 39.4 & 45.2 \\
\rowcolor{gray!15} \textbf{\quad + ANCHOR} & 41.2 & 34.8 & 47.5 & 39.0 & 53.2 & 37.7 & 33.9 & 40.5 & 46.7 & 56.6 \\

\midrule
\textbf{Models} & \cellcolor{zerobg}\textbf{om-ET} & \cellcolor{zerobg}\textbf{ro-RO} & \cellcolor{zerobg}\textbf{rw-RW} & \cellcolor{zerobg}\textbf{si-LK} & \cellcolor{zerobg}\textbf{su-ID} & \cellcolor{zerobg}\textbf{ta-IN} & \cellcolor{zerobg}\textbf{te-IN} & \cellcolor{zerobg}\textbf{ur-IN} & \cellcolor{zerobg}\textbf{ur-PK} & \textbf{Avg.} \\
\midrule
LLaVA-1.5-7B & 37.4 & 51.0 & 34.5 & 24.4 & 37.5 & 31.8 & 34.0 & 27.7 & 24.1 & 41.6 \\
\quad + Std. LoRA & 35.5 & 46.7 & 34.9 & 23.6 & 35.0 & 26.2 & 27.0 & 29.5 & 25.9 & 40.0 \\
\quad + Early-Layer FFT & 22.0 & 44.4 & 33.6 & 25.3 & 34.0 & 24.3 & 27.0 & 28.6 & 26.4 & 36.0 \\
\rowcolor{gray!15} \textbf{\quad + ANCHOR} & 34.9 & 37.4 & 35.3 & 24.1 & 36.4 & 29.0 & 33.2 & 32.3 & 26.6 & 42.0 \\
\midrule
LLaVA-1.5-13B  & 34.6 & 57.0 & 34.5 & 28.0 & 39.0 & 26.2 & 34.0 & 41.4 & 34.3 & 44.9 \\
\quad + Std. LoRA  & 32.2 & 52.3 & 36.6 & 30.2 & 38.0 & 32.2 & 31.5 & 37.7 & 34.7 & 43.1 \\
\quad + Early-Layer FFT    & 30.4 & 44.7 & 27.2 & 26.7 & 34.0 & 22.9 & 30.0 & 30.5 & 31.0 & 37.0 \\
\rowcolor{gray!15} \textbf{\quad + ANCHOR} & 32.0 & 58.0 & 37.1 & 30.9 & 38.7 & 33.3 & 32.6 & 39.9 & 38.3 & 45.9 \\
\bottomrule
\end{tabular}
}
\caption{CVQA benchmark performance. \colorbox{trainedbg}{Pink} columns denote language--region pairs whose languages are included in SFT, while \colorbox{zerobg}{Cyan} columns denote languages unseen during training. The \textbf{Avg.} column reports the macro-average accuracy across all 39 pairs.}
\label{tab:CVQA_local_full_reordered}
\end{table}

\subsection{Ablation Study}
\label{sec:ablation}
\paragraph{Target Layer ($L_{early}$)} 
The choice of $L_{early}$ is critical in our ANCHOR framework, as it defines the precise early translation window during which cross-modal alignment must occur. We conduct an ablation over different values of $L_{early}$ to understand its impact on downstream reasoning performance.

\Cref{tab:ablation_layer} presents the results of varying $L_{early}$ from 6 to 16 layers. We evaluate on xMMMU and MaXM, reporting the macro-averaged accuracy across all evaluated languages. 

The results reveal an inverted U-shaped trend: choosing $L_{early}$ too small (e.g., 6 layers) may provide insufficient time for visual semantics to emerge. Conversely, setting $L_{early}$ too large (e.g., 16 layers) may allow the linguistic trajectory to become strongly English-dominant before effective visual guidance is available. The best evaluated setting, $L_{early}=10$, is consistent with our mechanistic analysis of visual semantic emergence and achieves the highest downstream performance among the tested values. These results suggest that the timing of early-layer intervention is important for cross-modal alignment.

\paragraph{Intervention Scope: Early-Layer vs. Full SFT}
To further validate our layer-specific design, we compare our early-layer intervention ($L_{early}=10$) against a full-parameter fine-tuning baseline augmented with the PVA objective. Crucially, in this baseline, the PVA constraint remains applied at layer 10 to ensure a strictly controlled comparison. 
As shown in \Cref{tab:ablation_scope}, applying Full SFT degrades the overall accuracy across both benchmarks compared to our targeted early-layer approach. 
This performance gap suggests that full-depth updates may overfit to the SFT language distribution or interfere with representations that support cross-lingual generalization. 
Confining parameter updates to the early alignment window yields better aggregate performance in this comparison while preserving more of the LLM's multilingual capability.

\begin{table}[ht]
\centering
\begin{tabular}{lcc}
\toprule
\textbf{$L_{early}$} & \textbf{xMMMU} & \textbf{MaXM} \\
\midrule
6  & 32.8 & 28.5 \\
8  & 34.2 & 30.7 \\
\textbf{10} & \textbf{35.6} & \textbf{32.4} \\
12 & 34.7 & 31.8 \\
14 & 33.5 & 30.2 \\
16 & 32.1 & 28.9 \\
\bottomrule
\end{tabular}
\caption{Ablation study on the target layer $L_{early}$. xMMMU and MaXM results are macro-averaged across all evaluated languages. All experiments are conducted on LLaVA-1.5-7B.}
\label{tab:ablation_layer}
\end{table}

\begin{table}[ht]
\centering
\begin{tabular}{lcc}
\toprule
\textbf{Intervention Scope} & \textbf{xMMMU} & \textbf{MaXM} \\
\midrule
Full SFT + PVA & 32.8 & 27.5 \\
\textbf{ANCHOR (Ours)}   & \textbf{35.6} & \textbf{32.4} \\
\bottomrule
\end{tabular}
\caption{Ablation on intervention scope using LLaVA-1.5-7B. Full-parameter fine-tuning augmented with PVA at layer 10 yields lower aggregate performance than the early-layer intervention on both benchmarks, suggesting that targeted early-layer updates better preserve cross-lingual generalization in this setting.} 
\label{tab:ablation_scope}
\end{table}

\section{Complete Implementation Details}
\label{sup:implementation_details}
\paragraph{Model Architecture} The alignment network ($\text{MLP}_{\text{align}}$) consists of a two-layer perceptron with GELU activation.

\paragraph{Training Configuration} We train for 1 epoch on 135K image-text pairs. We use the AdamW optimizer with a peak learning rate of $2 \times 10^{-5}$ and a cosine decay schedule with a 3\% warmup period. The PVA loss weight is set to $\alpha=0.4$. All experiments are conducted on NVIDIA A100 (80GB) GPUs with a global batch size of 128. Input images are resized to $336 \times 336$ resolution, and the maximum sequence length is 2048 tokens. During training, we apply full-parameter fine-tuning to the multimodal projector and the first $L_{early}=10$ transformer layers, while freezing layers 11--32 (for 7B) or 11--40 (for 13B) and the vision encoder. The trainable parameters thus include the vision-language projector, the alignment network $\text{MLP}_{\text{align}}$, and the initial FFT transformer layers. For the standard LoRA baseline, we employ a rank $r=128$ and a scaling factor $\alpha_{\text{lora}}=256$, applying the adaptation to all linear layers (including $q, k, v, o$ and MLP gates) of the LLM backbone to ensure a competitive and fair comparison.

\end{document}